\pdfoutput=1
\documentclass[10pt,twocolumn,letterpaper]{article}

\usepackage{cvpr}
\usepackage{times}
\usepackage{epsfig}
\usepackage{graphicx}
\usepackage{amsmath}
\usepackage{amssymb}

\usepackage{hyperref}
\usepackage{hhline}
\usepackage{multirow}
\usepackage{algorithm}
\usepackage[noend]{algpseudocode}
\usepackage{subcaption}
\usepackage{epstopdf}
\usepackage{float}
\let\vec\mathbf

\cvprfinalcopy 

\def\cvprPaperID{***} 

\ifcvprfinal\fi
\begin{document}

\title{Richer and Deeper Supervision Network for Salient Object Detection}

\author{Sen Jia \\
Ryerson University\\
{\tt\small sen.jia@ryerson.ca}
\and 
Neil D. B. Bruce\\
Ryerson University, Vector Institute\\
{\tt\small bruce@ryerson.ca}
}

\maketitle

\begin{abstract}

Recent Salient Object Detection (SOD) systems are mostly based on Convolutional Neural Networks (CNNs). Specifically, Deeply Supervised Saliency (DSS) system has shown it is very useful to add short connections to the network and supervising on the side output. In this work, we propose a new SOD system which aims at designing a more efficient and effective way to pass back global information. Richer and Deeper Supervision (RDS) is applied to better combine features from each side output without demanding much extra computational space. Meanwhile, the backbone network used for SOD is normally pre-trained on the object classification dataset, ImageNet. But the pre-trained model has been trained on cropped images in order to only focus on distinguishing features within the region of the object. But the ignored background information is also significant in the task of SOD. We try to solve this problem by introducing the training data designed for object detection. A coarse global information is learned based on an entire image with its bounding box before training on the SOD dataset. The large-scale of object images can slightly improve the performance of SOD. Our experiment shows the proposed RDS network achieves the state-of-the-art results on five public SOD datasets.

\end{abstract}

\section{Introduction}

Human beings tend to focus more on salient regions within a natural image. Using machines to predict where humans fixate on, known as saliency prediction, plays an important role in various vision tasks, e.g. image classification \cite{Sharma12,Lei15}, visual tracking \cite{Mahadevan09,Borji12} and semantic segmentation \cite{Hou16, Jin17}. Most of the recently proposed Salient Object Detection (SOD) systems are based on deep Convolutional Neural Networks (CNNs) to extract informative features \cite{MDF,DCL,ELD,DS,NLDF,UCF,AMU,SRM,CAR,DSS,DGRL,PIC,LPSD,PAGR,CKT}. It has been proven that the depth of CNN models affects the performance of the learned feature, e.g. object recognition \cite{VGG, ResNet}, fixation prediction \cite{Cornia18,EML} or face classification \cite{Sun15,Jia16}. Intuitively, a deeper CNN model can also better extract low-level and high-level features for SOD from the bottom and the top side outputs respectively. 

\begin{figure}[t]
\hspace*{-1.5em}
\begin{subfigure}{1.1\columnwidth}
\centering
\includegraphics[width=1\columnwidth]{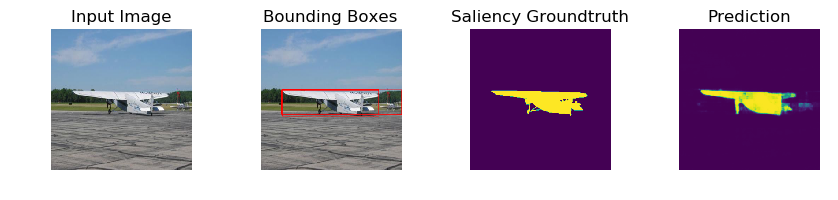}
\end{subfigure}

\hspace*{-1.4em}
\begin{subfigure}{1.1\columnwidth}
\centering
\includegraphics[width=1\columnwidth]{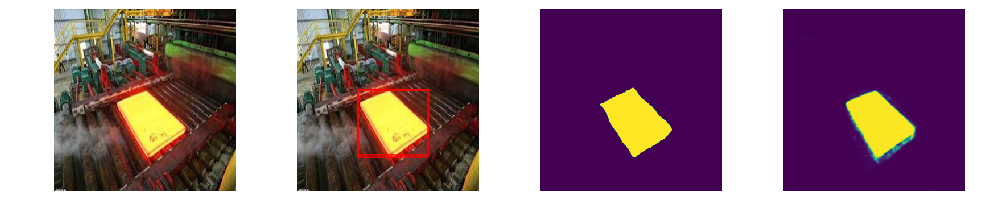}
\end{subfigure}

\hspace*{-1.5em}
\begin{subfigure}{1.1\columnwidth}
\centering
\includegraphics[width=1\columnwidth]{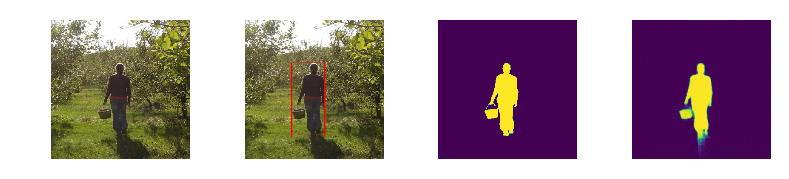}
\end{subfigure}

\caption{Examples of object detection (bounding boxes) and salient object detection (pixel-wise annotations).}
\label{fig:example1}
\end{figure}

In practice, a deep CNN architecture may not be able to deliver a better performance if the combination of global and local features is not well-designed. Moreover, it is difficult to stack more layers for SOD because the saliency-oriented designs require much computational space e.g. recurrent module \cite{PIC,PAGR}, stage-wise refinement \cite{SRM,CAR} or side outputs \cite{DSS,NLDF}. Another problem in the use of deep CNNs is the limited training data for SOD. Comparing with bounding boxes for object detection, the available data for SOD is much smaller due to the labour-intensive process of pixel-level labelling. In this paper, we intend to solve these two problems by proposing a new network structure and introducing related annotated data from the vision task of object detection.

\begin{figure*}[!hpt]
\begin{center}
\begin{subfigure}{0.65\columnwidth}
\centering
\includegraphics[width=1\columnwidth]{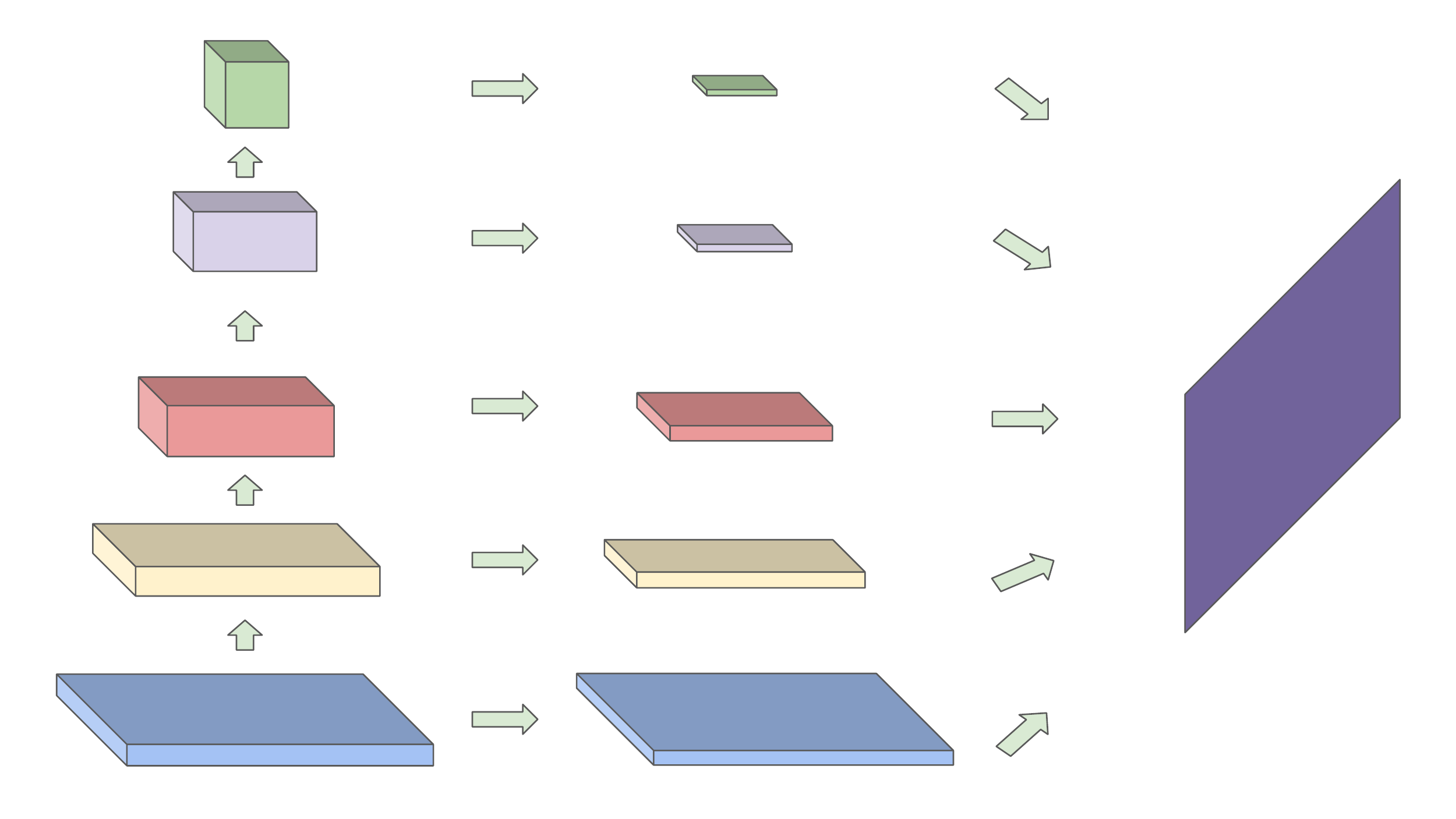}
\caption{}\label{fig:hedflow}
\end{subfigure}
\begin{subfigure}{0.65\columnwidth}
\centering
\includegraphics[width=1\columnwidth]{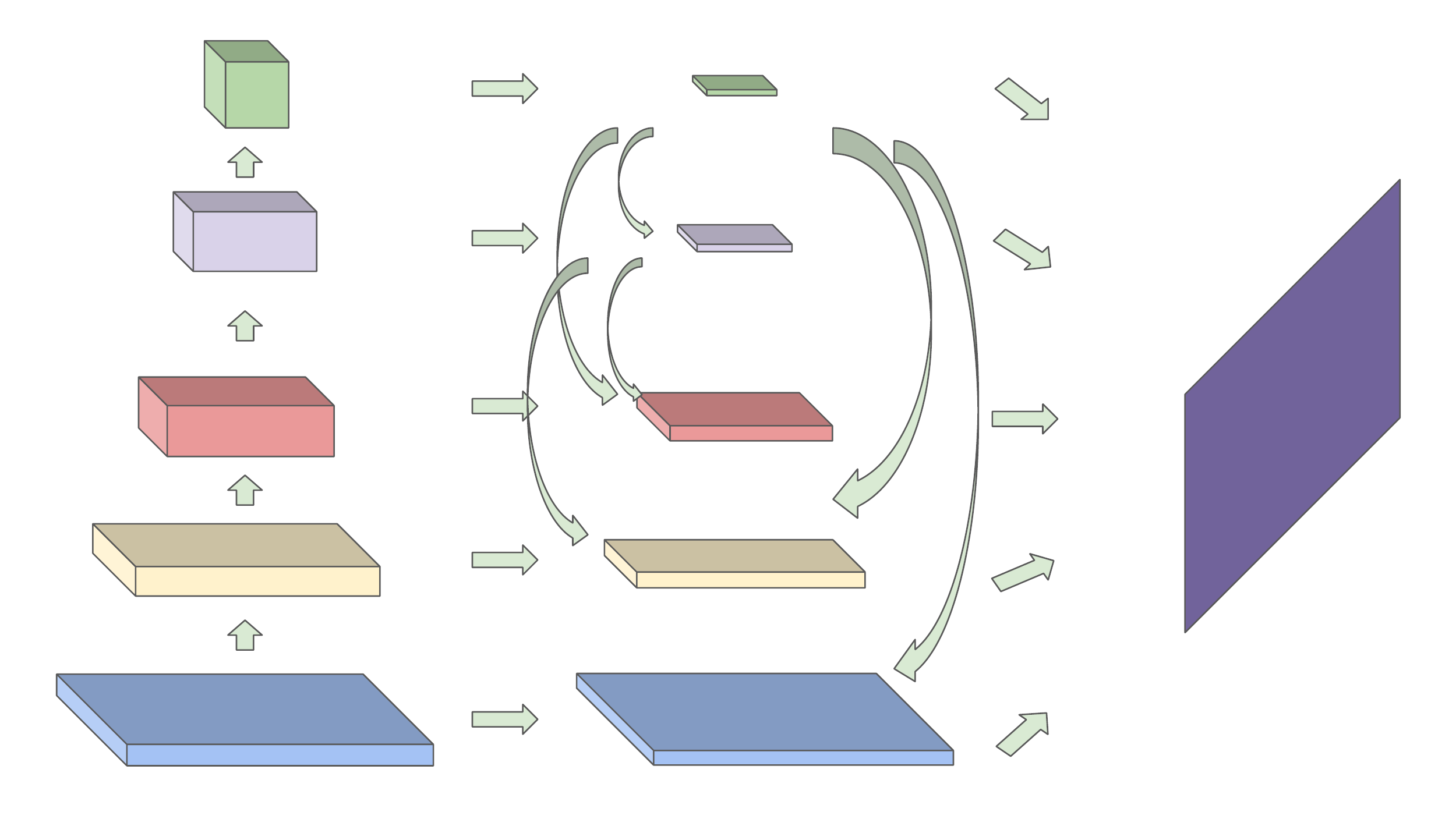}
\caption{}\label{fig:dssflow}
\end{subfigure}
\begin{subfigure}{0.65\columnwidth}
\centering
\includegraphics[width=1\columnwidth]{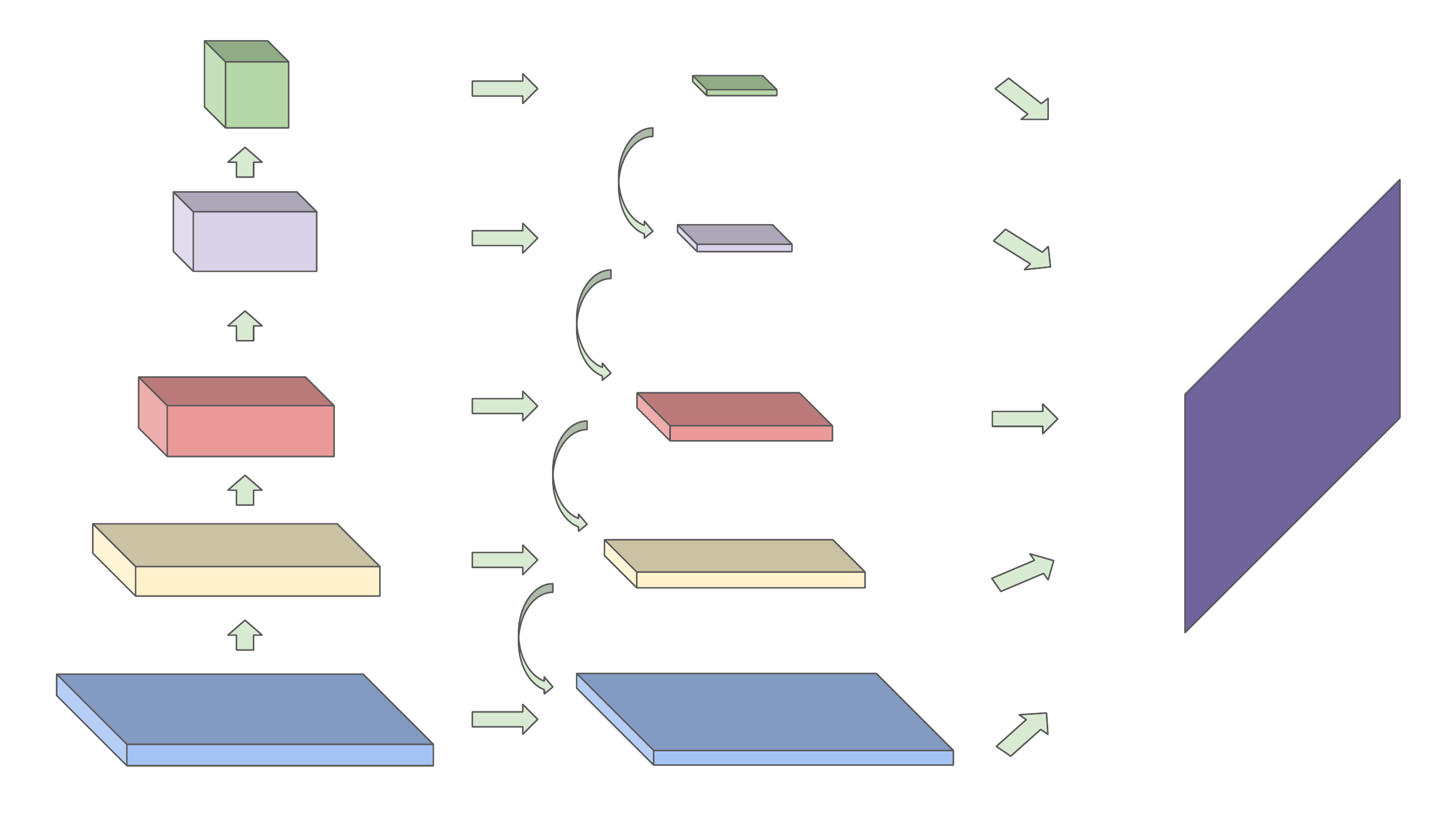}
\caption{}\label{fig:rdsflow}
\end{subfigure}
\end{center}
   \caption{Flowchart of the networks. (a) HED \cite{HED}. (b) DSS \cite{DSS}. (c) RDS.}
   \label{fig:flow}
\end{figure*}

It has been shown to be effective to apply supervision on side outputs of a CNN model for contour detection \cite{HED}. Furthermore, this idea has been applied successfully for SOD using short connections (DSS) \cite{DSS}. But the design of DSS passes limited global information to bottom layers, which may hinder the performance. Increasing the number of channels will require a huge number of parameters for the transposed convolution. In this study, we propose a new network aiming at delivering richer representations from deeper CNN layers, denoted as RDS network. More global and local features are retained and combined in a more efficient way, see Section~\ref{sec:rds}.

To overcome the bottleneck of limited data, one solution is learning SOD features from closely related vision tasks, e.g. using bounding boxes \cite{Jia13} or contour knowledge \cite{CKT}. It is common to fine-tune a CNN model that pre-trained on the object dataset, ImageNet \cite{ImageNet}, to extract saliency features. This suggests the learned feature from one vision task can be used for another. Furthermore, a recent survey \cite{Survey} has shown the close relationship among SOD, fixation prediction and object detection, see Figure~\ref{fig:example1}. There also exist studies for object detection based on saliency features \cite{Alexe10, Siva13}. Therefore, we propose to learn SOD features using the annotated data which is designed for object detection. The advantages of using object data are two-fold: a) the number of available annotated images is much larger than SOD because bounding box is easier to label. b) comparing with object classification, the images contain back ground information which is informative to SOD. The ImageNet DET \cite{ImageNet} and the PASCAL-VOC \cite{PASCALVOC} datasets are merged to formulate a new dataset to pre-train our RDS network for a coarse knowledge about location, see Section~\ref{sec:obj}. 

In our experiment, we directly compare our RDS with the DSS network under the same setting to show that RDS can combine global and local features more effectively. We can have a slight gain for SOD after training on the merged object dataset. Finally, we compare our RDS network with other state-of-the-art methods on five public SOD datasets, HKU-IS \cite{MDF}, ECSSD \cite{ECSSD}, PASCAL-S \cite{PASCALS}, DUT-OMRON \cite{OMRON} and DUT-S \cite{DUTS}. The experiment shows our method outperforms most of the competitors on the metrics of F-measure and Mean Absolute Error (MAE).

\section{Related Work}
\label{sec:related}

\begin{figure*}[pht]
\begin{center}
\includegraphics[width=2\columnwidth]{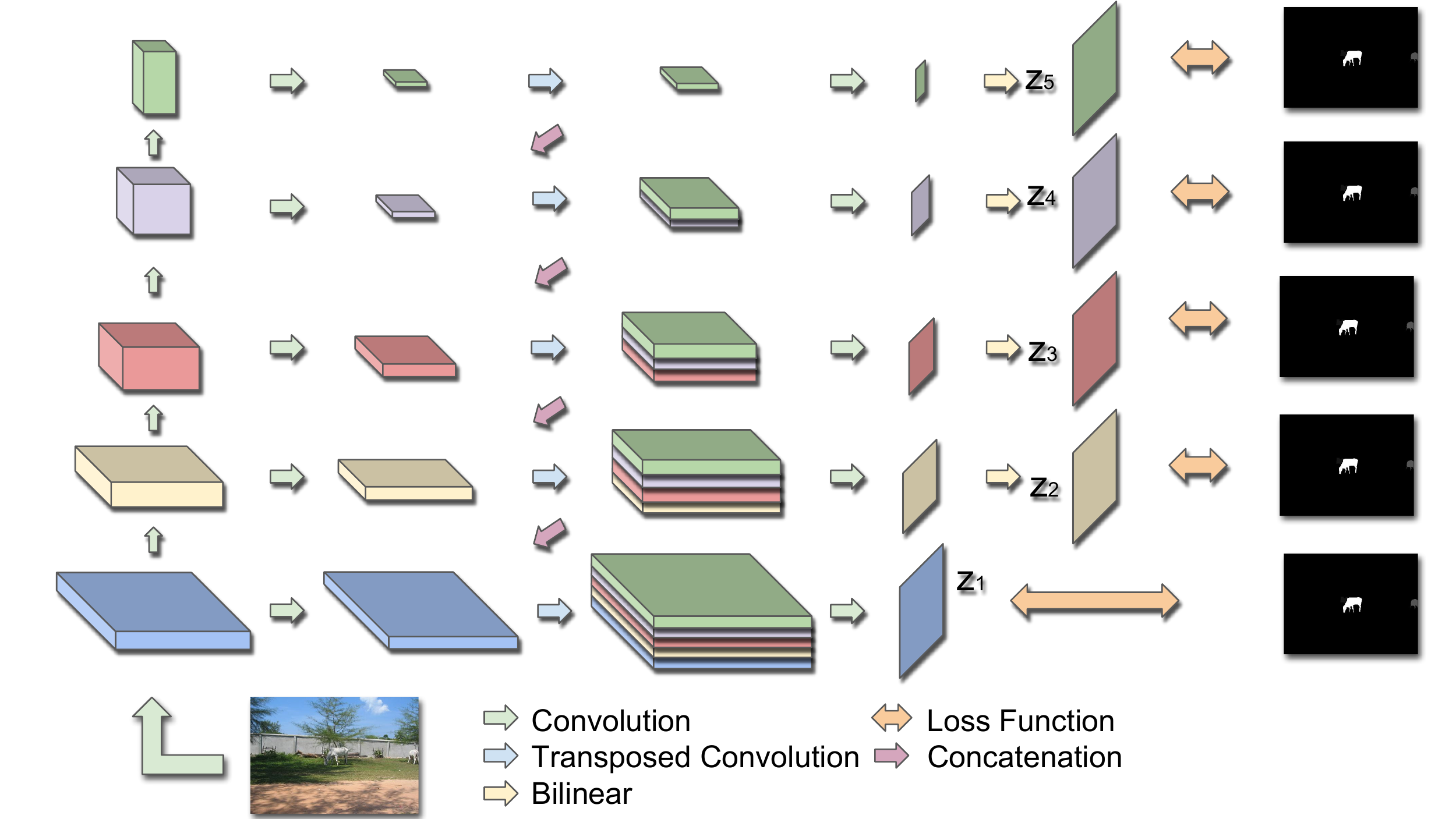}
\end{center}
\caption{Flowchart of our proposed RDS network.}
\label{fig:RDS}
\end{figure*}

Early SOD methods are mainly based on hand-crafted features to capture saliency information, e.g. global contrast \cite{Cheng15}, multiple color spaces \cite{Kim14}, clustering on reconstruction error \cite{Li13} and objectness cues \cite{Jia13}. But traditional methods can hardly outperform CNN-based because low-level features are not rich enough to encode saliency information. In this section, we mainly discuss CNN-based state-of-the-art methods because they are more related to our work.

Li \textit{et al.} \cite{MDF} proposed to apply CNN on three different scales of segmentation. The extracted information is merged to compute the final saliency map. Later they proposed a two-stream system to combine pixel-wise and segment-wise features \cite{DCL}. Lee \textit{et al.} generated a distance map between a query region and other parts, and the map is combined with CNN-based features for SOD \cite{ELD}. Li \textit{et al.} \cite{DS} applied regularized regression on super-pixel to shape CNN features for a more precise boundary. Luo \textit{et al.} \cite{NLDF} combined local and global information using multiple side outputs and the Mumford-Shah functional was also applied to measure smoothness within each sub-regions. Zhang \textit{et al.} \cite{UCF} proposed reformulated dropout (R-dropout) to formulate uncertain ensemble CNN features. The R-dropout operation can be drawn from any probability distribution but the summation of mask weights is one. The authors also proposed to combine coarse semantics and details at multi-level resolutions \cite{AMU}. Wang \textit{et al.} added a refinement module on the output of a CNN model. The refinement module consists of another CNN model and CNN filters with different sizes and strides for pyramid pooling. Amirul \textit{et al.} \cite{CAR} also applied a refinement module to shape a sharper boundary but their method is based on large kernel filters to capture global context. Zeng \textit{et al.} \cite{LPSD} applied a CNN model to map pixels and regions of an image into a sub-space, in which nearest neighbour is utilized to find salient regions. Wang \textit{et al.} \cite{DGRL} applied recurrent network on intermediate CNN features from ResNet for global localization. Another network is applied to refine the boundary by multiplying a coefficient. Liu \textit{et al.} \cite{PIC} used the ReNet \cite{ReNet} model, a recurrent network, to capture global context. The local feature is extracted by using CNN filters with large kernel. Zhang \textit{et al.} \cite{PAGR} applied recurrent network to integrate attention information in a progressive way. Softmax and average pooling are applied on each channel to create channel-wise attention, which is later being element-wise multiplied by the traditional spatial activation. Li \textit{et al.} \cite{CKT} proposed a system that can learn contour and SOD features at the same time to further improve the performance. 

One recently proposed method \cite{DSS} is closely related to ours because both methods exploit multiple side outputs and short connections. As shown in Figure~\ref{fig:flow}, HED (\ref{fig:hedflow}) was proposed to detect contour, but no short connections are applied. The DSS network (\ref{fig:dssflow}) connects each layer to all the previous layers. But the large-size transposed convolution requires much space to compute, it becomes more problematic when passing more channels. While our design of RDS (\ref{fig:rdsflow}) only concatenates neighbouring side outputs using transposed convolution with a small kernel to save space. 

Another related work is \cite{Jia13} because we both utilize the objectness cue to predict salient regions without using category information. Their method \cite{Jia13} used various hand-crafted features to propose box candidates. Pixel-level scores are computed by applying the Gaussian process. Our method is CNN-based and we try to simply convert the task of object detection into SOD.

\section{Methodology}

\begin{table*}[t]
\begin{center}
\begin{tabular}{|c|c c c|c|}
\hline
Side Output & Layer 1 & Layer 2 & Layer 3 & Channels\\
\hhline{|=|=|=|=|=|}
\textit{conv1} &Conv3@128&Conv3@128&Conv1@32&(160, H, W)\\ 
\textit{res2c} &Conv5@256&Conv5@256&Conv1@32&(128, H/2, W/2)\\ 
\textit{res3d} &Conv5@256&Conv5@256&Conv1@32&(96, H/4, W/4)\\ 
\textit{res4f} &Conv5@512&Conv5@512&Conv1@32&(64, H/8, W/8)\\ 
\textit{res5c} &Conv7@512&Conv7@512&Conv1@32&(32, H/16, W/16)\\ 
\hline
\end{tabular}
\end{center}
\caption{Table of convolutional filters for each side output.}
\label{tab:filters}
\end{table*}

Let's first define the input data as $(\vec{x_i}, \vec{y_i}), i=1,2 \dots N$, where $\vec{x_i}$ and $\vec{y_i}$ represent the $i$th input image and its ground truth respectively, totally $N$ images in the training set. For each input image, $\vec{x} \in R^{H \times W \times C}$, a transformation function $f$ (a series of convolutional operations) is applied to predict a saliency map, $\vec{\hat{y}} = f(\vec{\theta}, \vec{x}) \in R^{H \times W}$, where $\vec{\theta}$ is the weight of the model. Let $Z=\{\vec{z}_1, \vec{z}_2, \dots, \vec{z}_M, \vec{z}_{fuse}\}$ denote predicted maps from different outputs and the final fused one, where $M$ is the total number of side outputs. We apply ResNet \cite{ResNet} as our backbone in this paper, five CNN layers ($M=5$) are selected as side outputs, \textit{conv1}, \textit{res2c}, \textit{res3d}, \textit{res4f} and \textit{res5c}. The loss for $m$th side output can be computed:

\begin{equation}
\begin{split}
&\ell_m = \sigma(\vec{z}_{m}, \vec{y}), \quad \vec{z}_{m}=f(\vec{x}, \vec{\theta}) \\
&\vec{z}_{fuse} = h(\sum_{m=1}^{M}{\vec{z}_{m}}) \\
&L_{total} = \ell_{fuse} + \sum_{m=1}^{M}{\ell_m}
\end{split}
\end{equation}\label{eq:loss}

where $\sigma$ denotes the loss metric, we use mean squared loss in our experiment. The total loss to optimize is the summation of each side loss and the fused one.

\subsection{Richer and Deeper Supervision}
\label{sec:rds}

Supervising on CNN side outputs was applied for edge detection \cite{HED} by extracting five layers from the VGGNet \cite{VGG}. Later the architecture was extended for SOD \cite{DSS} by connecting one more CNN layer from \textit{pool5} and two additional convolutional layers were added on each side output. In their study, maps from higher levels contain more information about location whereas lower side outputs pay more attention on details. But the design of DSS compresses each side output into one channel before concatenating with previous layers and we believe there is a significant information loss in the short connection.

To leverage the power of deep CNN models, we encode each side output into $k$ ($k=32$) channels to retain more features, see Table~\ref{tab:filters}. But this ``wider'' short connection requires more space for computation such that it is difficult to deploy a ``deeper'' CNN model. For instance, there are $64 \times 64 = 4,096$ parameters when applying transposed convolution between the top and the bottom outputs. It grows linearly $4,096 \times k$ when increasing the number channels. We can not deploy this design (ResNet152 with $32$ channels using DSS) on one Nvidia Titan graphic card, even though dropout is applied on the short connections according to Equation($19$) in \cite{DSS}.

To solve the trade-off between ``wide'' and ``deep'', we avoid direct connection from the top to the bottom as used in Figure~\ref{fig:dssflow}. Large-size transposed convolutional filters have been removed in our design. Instead, we apply small filters, two by two, with a stride of two for up-sampling. Each side output only connects to the previous one layer. The last column of Table~\ref{tab:filters} shows the number of concatenated channels and the size of each output after up-sampling. The bottom side output has $160$ channels, that is the number of required parameters for up-sampling is $640$, five time less than the large size filter used in DSS. In this way, a richer representation of global view can be merged with low-level features. Although more computational load is needed between \textit{Layer 2} and \textit{Layer 3} in Table\ref{tab:filters}, our RDS with ResNet-152 can be trained using a single Titan card. Batch normalization and ReLU are added after each convolution and transposed. Figure~\ref{fig:RDS} shows the whole RDS network system and the final output is the fused saliency map from Equation~\ref{eq:loss}. 

\subsection{Category-Independent Objectness Cues}
\label{sec:obj}

\begin{table*}[pt]
\begin{center}
\begin{tabular}{|c|c|c|c|c|c|c|c|c|c|c|c|}
\hline
\multirow{2}{*}{Methods} & \multicolumn{2}{c|}{ECSSD}& \multicolumn{2}{c|}{HKU-IS} & \multicolumn{2}{c|}{PASCAL-S}& \multicolumn{2}{c|}{DUT-OMRON}& \multicolumn{2}{c|}{DUTS-TE} \\ 
  &$F_{\beta}$& $MAE$ &$F_{\beta}$& $MAE$ &$F_{\beta}$& $MAE$ &$F_{\beta}$& $MAE$ &$F_{\beta}$& $MAE$\\ 
 \hhline{|=|=|=|=|=|=|=|=|=|=|=|}
 DSS-50  &.944&.042&.933&.033&.868&.086&.825&.052&.860&.047\\ 
  DSS-152  &.947&.038&.937&.031&.871&.083&.837&.048&.866&.044\\ 
 \hline
 RDS-50  &.948&.037&.939&.029&.874&.082&.824&.054&.863&.044\\
 RDS-152  &.953&.036&.942&.028&.874&.080&.837&.050&.867&.044\\ 
\hline
\end{tabular}
\end{center}
\caption{Comparison between RDS and DSS based on ResNet-50 and ResNet-152.}
\label{tab:rds_dss}
\end{table*}

The annotation for SOD is very difficult because it requires precise pixel-level labelling. While it is easier to obtain labelled training data for object classification or detection because only category labels or bounding boxes are needed respectively. To further exploit our rich and deep design, object images for detection (bounding boxes) are used to learn coarse SOD features.

\begin{figure}[!pth]
\begin{center}
\includegraphics[width=1\columnwidth]{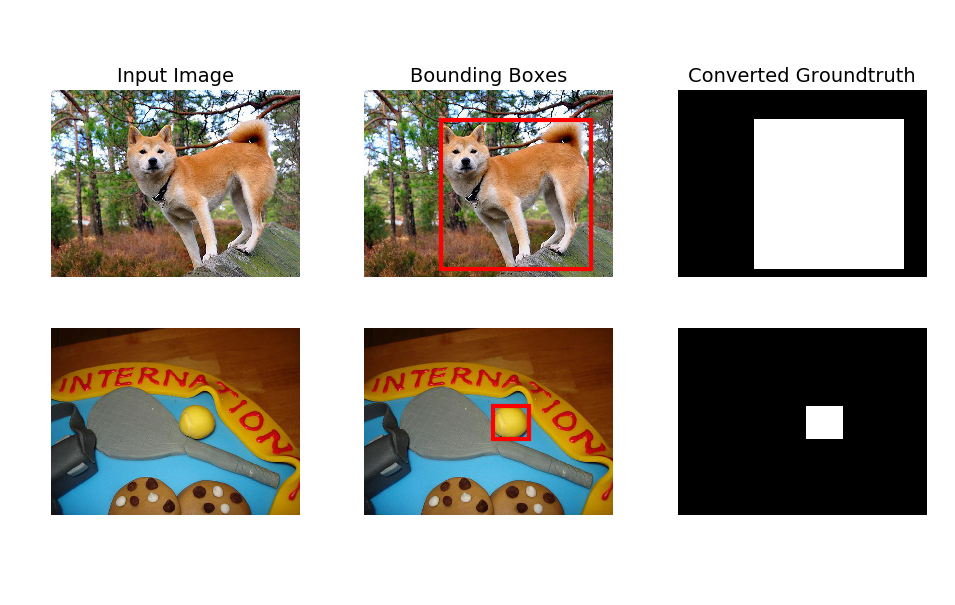}
\end{center}
   \caption{Examples from the object detection datasets with bounding boxes .}
   \label{fig:obj}
\end{figure}

The CNN model pre-trained on ImageNet \cite{ImageNet} contains knowledge about objects. But the models, VGG \cite{VGG} and ResNet \cite{ResNet}, only focus on distinguishing features of some specific objects. The location of the object or noisy background is ignored by center cropping and this leads to contextual information loss. Therefore, we introduce the datasets designed for object detection, VOC (2010) \cite{PASCALVOC} and ImageNet \cite{ImageNet}, to learn objectness cues, as shown in Figure~\ref{fig:obj}. 

The VOC dataset contains $11,321$ images with bounding boxes and ImageNet \cite{ImageNet} has $369,400$. These two datasets are much larger than the largest available SOD dataset, DUTS \cite{DUTS}. Therefore, we merged the two datasets to compile a new dataset, denoted as object dataset to distinguish from saliency datasets. Only the training sets were merged because we report our saliency result on the PASCAL-S \cite{PASCALS} and the DUTS \cite{DUTS} datasets. The bounding box of each image was converted to a saliency ground truth. Given an image $\vec{x}$ from the object dataset, a new saliency is initialized by $\vec{y}_{obj} \gets \vec{x}, \vec{y}_{obj} \in R^{H \times W}$. Then salient region within the map is computed by:

\begin{align}
\vec{y}_{obj}(i,j)&=\begin{cases}
    1, & \text{if $(i,j)$ within a bounding box}.\\
    0, & \text{otherwise}.
  \end{cases}\\
  &i=1\dots H, \quad j=1\dots W \nonumber
\end{align}

The resulting binary saliency map $\vec{y}_{obj}$ is the ground truth of the object dataset, as shown in Figure~\ref{fig:obj}. The map can be considered as a coarse salient region without pixel-level annotations. We try to further refine this dataset by removing some uninformative maps with low entropy. We formulate this problem by simply computing:
\begin{equation}
    \alpha = \frac{1}{H \times W}\sum_{i=1}^{H}{\sum_{j=1}^{W}{\vec{y}_{obj}(i,j)}}
\end{equation}

We only kept those images with $\alpha < 0.8$ to train the RDS network. Lower $\alpha$ value, e.g. less than $0.2$, is also considered as containing rich background information. Finally, the total number of the object dataset contains $305,401$ images, $4,217$ from VOC and $301,184$ from ImageNet respectively. 

\section{Experimental Setup}
\subsection{Saliency Data sets}
Besides the object dataset, six public saliency datasets are used for training and testing.

\paragraph{SOD Training Set}:
Two large-scale SOD datasets are used to train our RDS network. THUS10K (THUS) and DUT-S. The THUS dataset \cite{MSRAK} contains $10,000$ annotated images. The data was collected from the MSRA dataset which is for object detection \cite{MSRA-Liu}. The whole THUS dataset is used to learn SOD features. The DUT-S dataset \cite{DUTS} contains $10,533$ images for training and $5,019$ for testing. Only the training set of DUT-S, (DUTS-TR), is used for training. Totally we have $20,533$ images in our SOD training set.

\paragraph{SOD Test Set}: The $5,019$ images from the DUT-S test set are used to report result, denoted as DUTS-TE \cite{DUTS}. The ECSSD \cite{ECSSD} dataset contains $1,000$ images which are collected from internet and annotated by five helpers. HKU-IS \cite{MDF} has $4,447$ annotated images which are considered to be more challenging than THUS, because more disconnected objects or objects near the boundary are included. The PASCAL-S dataset \cite{PASCALS} contains $850$ pixel-wise labelled images from the test set of PASCAL-VOC 2010 challenge \cite{PASCALVOC}. DUT-OMRON \cite{OMRON} was built based on the SUN dataset \cite{SUN}, more than five thousand images are annotated for SOD. These five datasets are used to report result and compare with other methods.

\subsection{Evaluation Metrics}
We evaluate our RDS network using three measures, Precision-Recall (PR) curve, F-measure score and Mean Absolute Error (MAE). The output saliency map, $\vec{Z}_{fuse}$, is continuous and in the range of $[0, 255]$. The normalized saliency map and ground truth are re-scaled to $[0, 1]$, denoted as $\tilde{\vec{Z}}_{fuse}$ and $\tilde{\vec{y}}$. The MAE can be computed by:

\begin{equation}
    MAE = \frac{1}{H \times W}\sum_{i=1}^{H}{\sum_{j=1}^{W}{|\tilde{\vec{Z}}_{fuse} - \tilde{\vec{y}}|}}
\end{equation}

The output saliency map is also converted into a binary map by applying different threshold values. This binarized map is used to compute PR curve as well as the F-measure with a coefficient $\beta$:

\begin{equation}
    F_{\beta}=\frac{(1+\beta^2)Precision \times Recall}{(\beta^2Precision) + Recall}
\end{equation}
We choose $\beta^2=0.3$ as suggested by \cite{Achanta09} to pay more attention on precision and the maximum F-measure is reported in our experiment.

\subsection{Augmentation and CRF}
During each training epoch, we augment the training set by horizontally flipping the input image with a probability of $0.5$. After generating the final output, we apply the fully connected Conditional Random Field (CRF) \cite{crf} on each map as a post process, implemented within PyDenseCRF \footnote{\url{https://github.com/lucasb-eyer/pydensecrf}}.

\subsection{Validation of RDS Network}
\label{sec:exp}

We first compare the proposed RDS network with DSS under the same training protocol, the merged object dataset is not applied in this experiment. The DSS and RDS are applied on both ResNet-50 and ResNet-152 to show the differences of depth of the model and the design of connections. The final output of RDS is the fused one, $z_{fuse}$. But DSS predicts the saliency map using the average of the selected outputs \cite{DSS}, $\vec{\hat{y}} = Mean(\vec{z_2, z_3, z_4, z_{fuse}})$ as used in \cite{DSS}. 

The input image was resized to $320$ by $320$ and the number of training epoch was $25$. The initial learning rate was set to $0.01$ and it was reduced by multiplying a coefficient of $0.1$ every $10$ epochs. Weight decay was set to $1e^{-4}$ and the momentum value was $0.9$. The size of the training batch was $8$. The whole network was trained using a single Nvidia Titan graphic card. 

\begin{figure}[pht]
\begin{center}
\includegraphics[width=1\columnwidth,height=0.45\columnwidth]{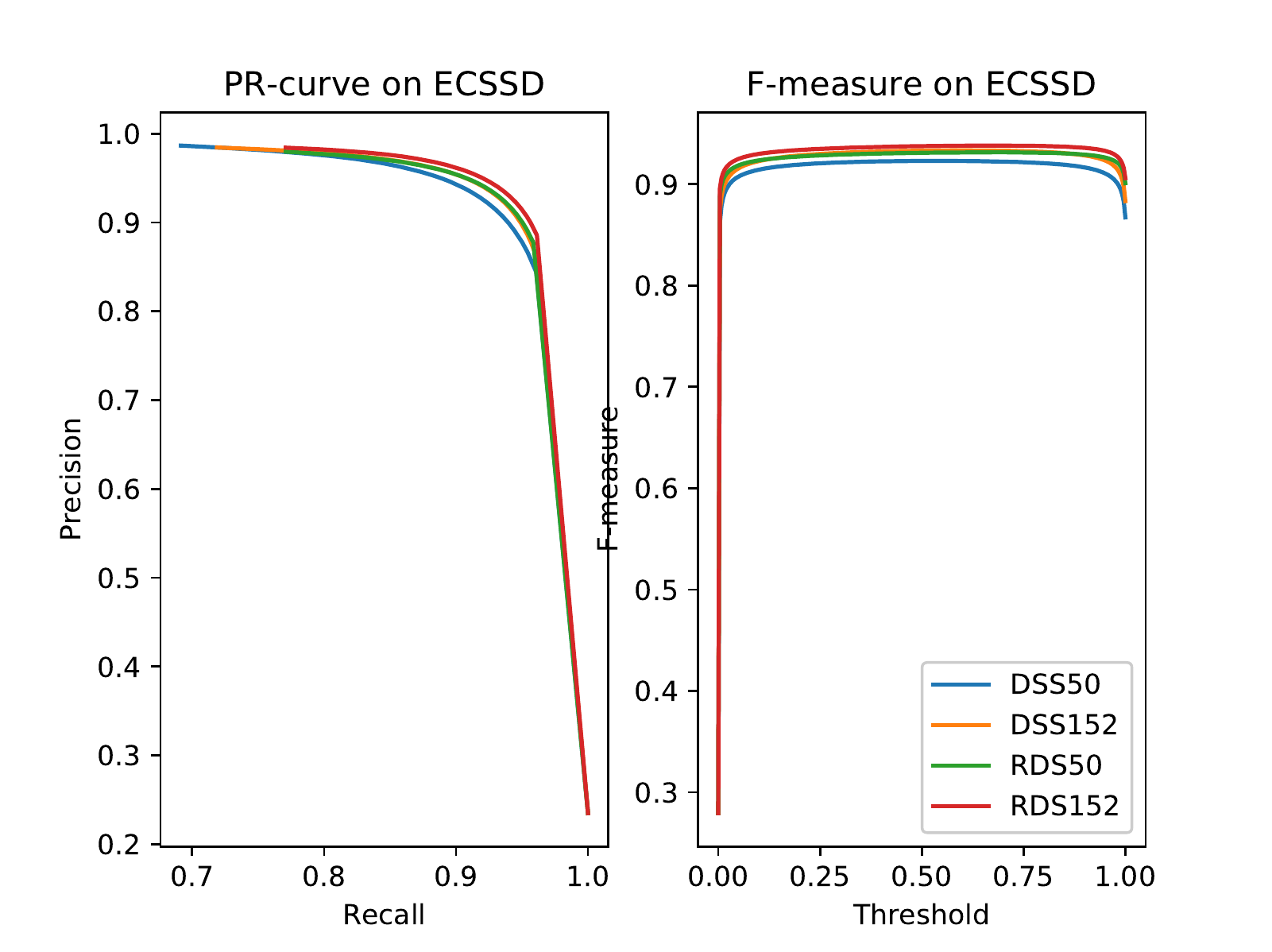}
\includegraphics[width=1\columnwidth,height=0.45\columnwidth]{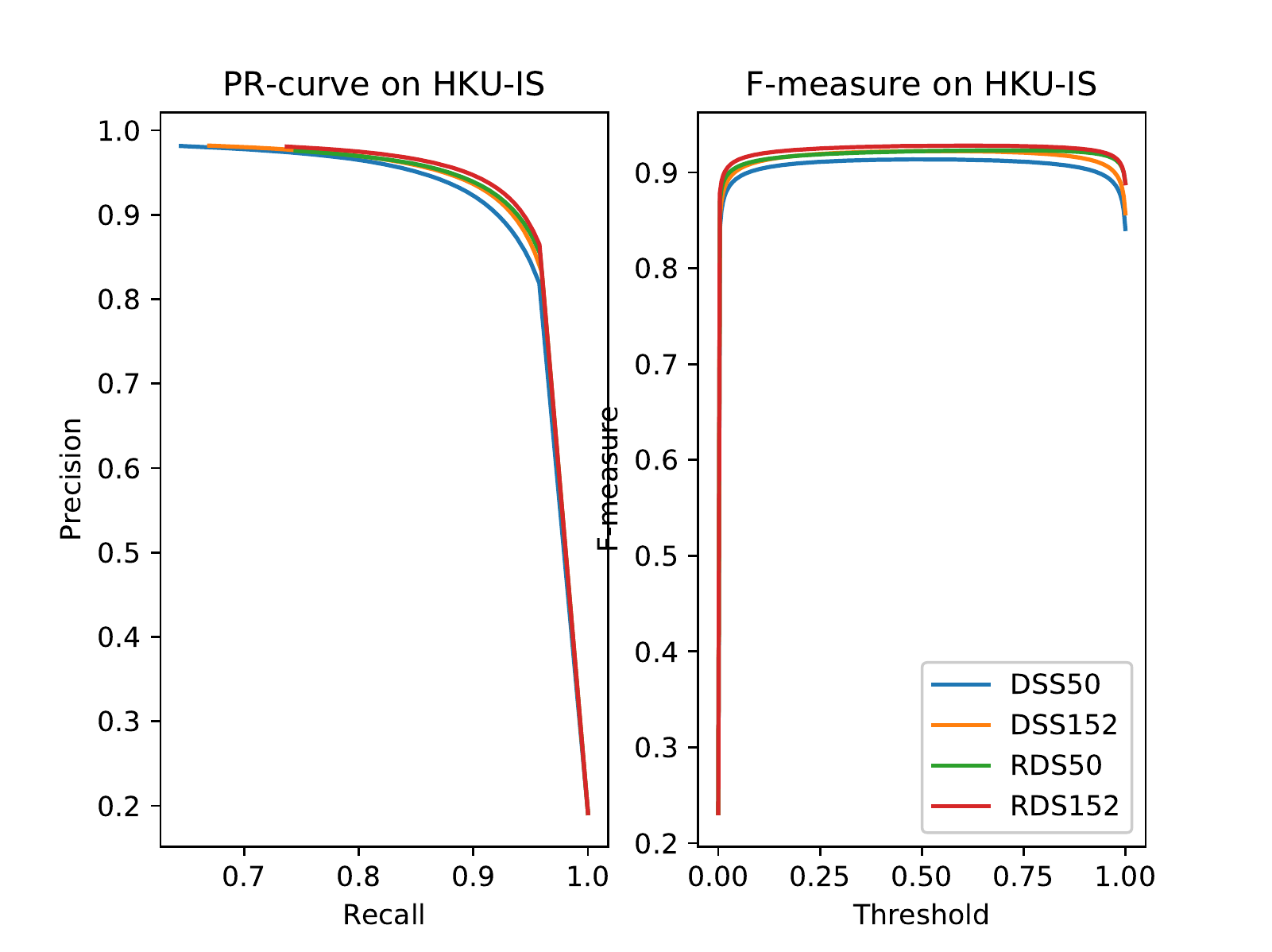}
\includegraphics[width=1\columnwidth,height=0.45\columnwidth]{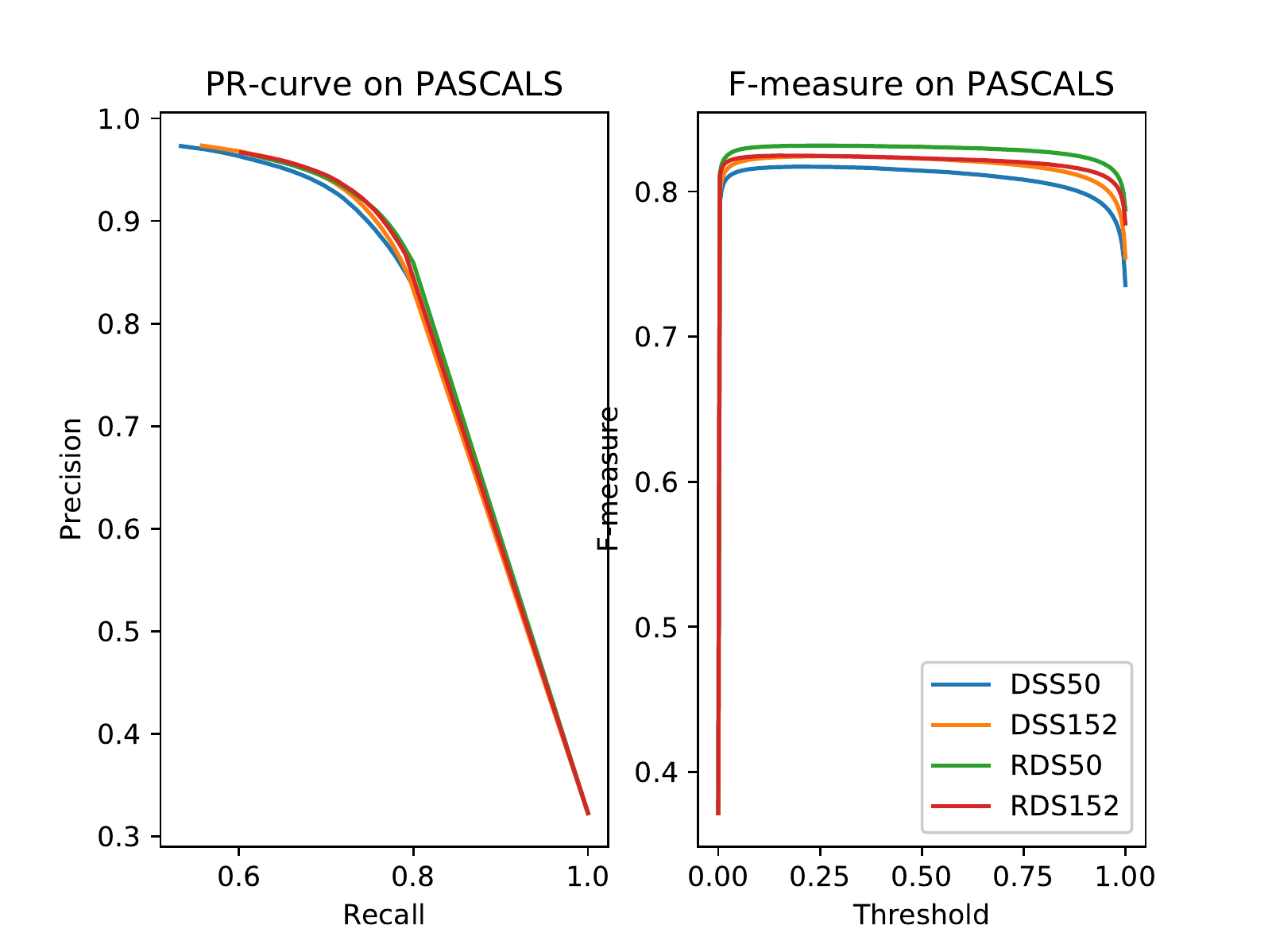}
\includegraphics[width=1\columnwidth,height=0.45\columnwidth]{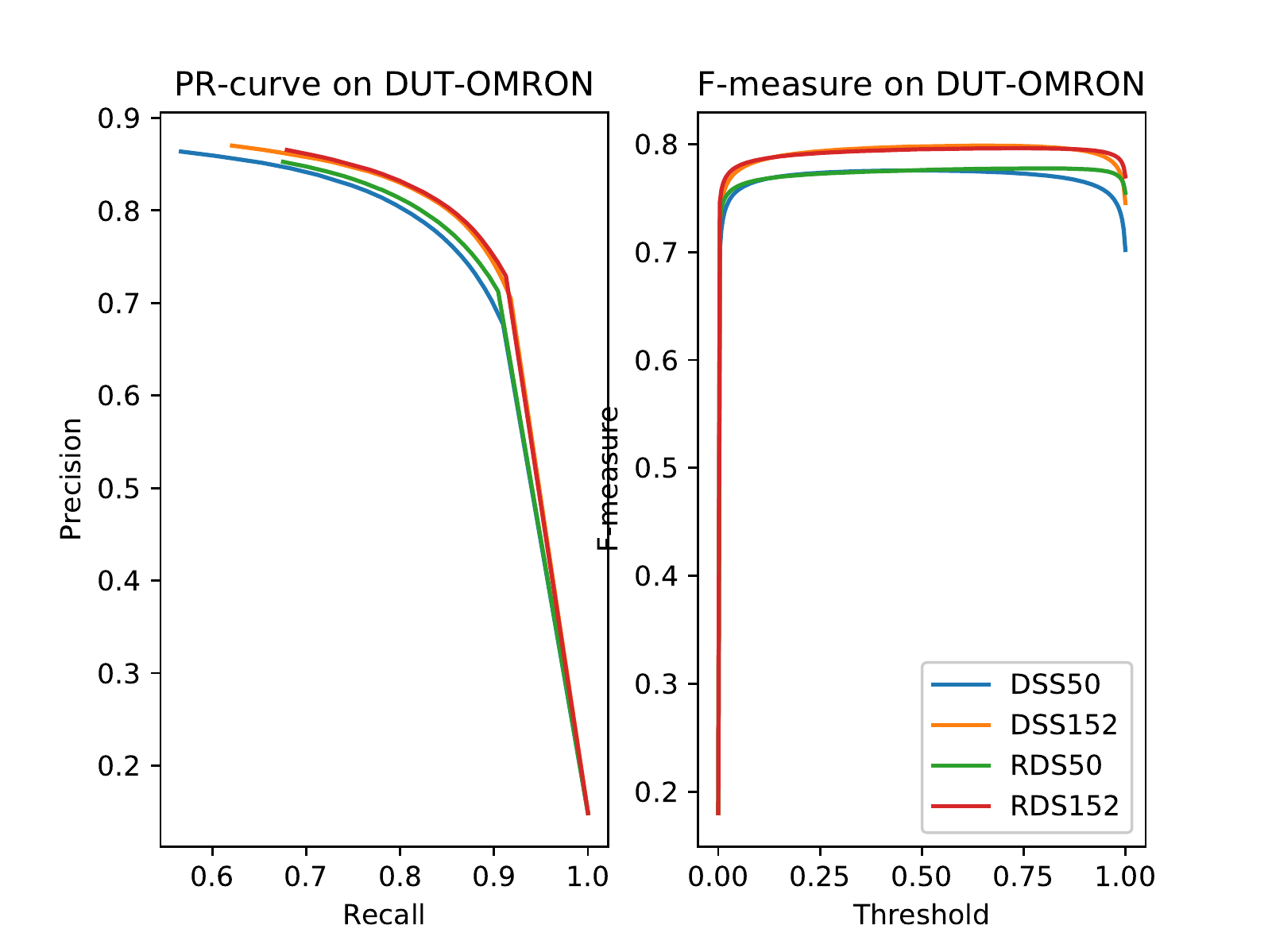}
\includegraphics[width=1\columnwidth,height=0.45\columnwidth]{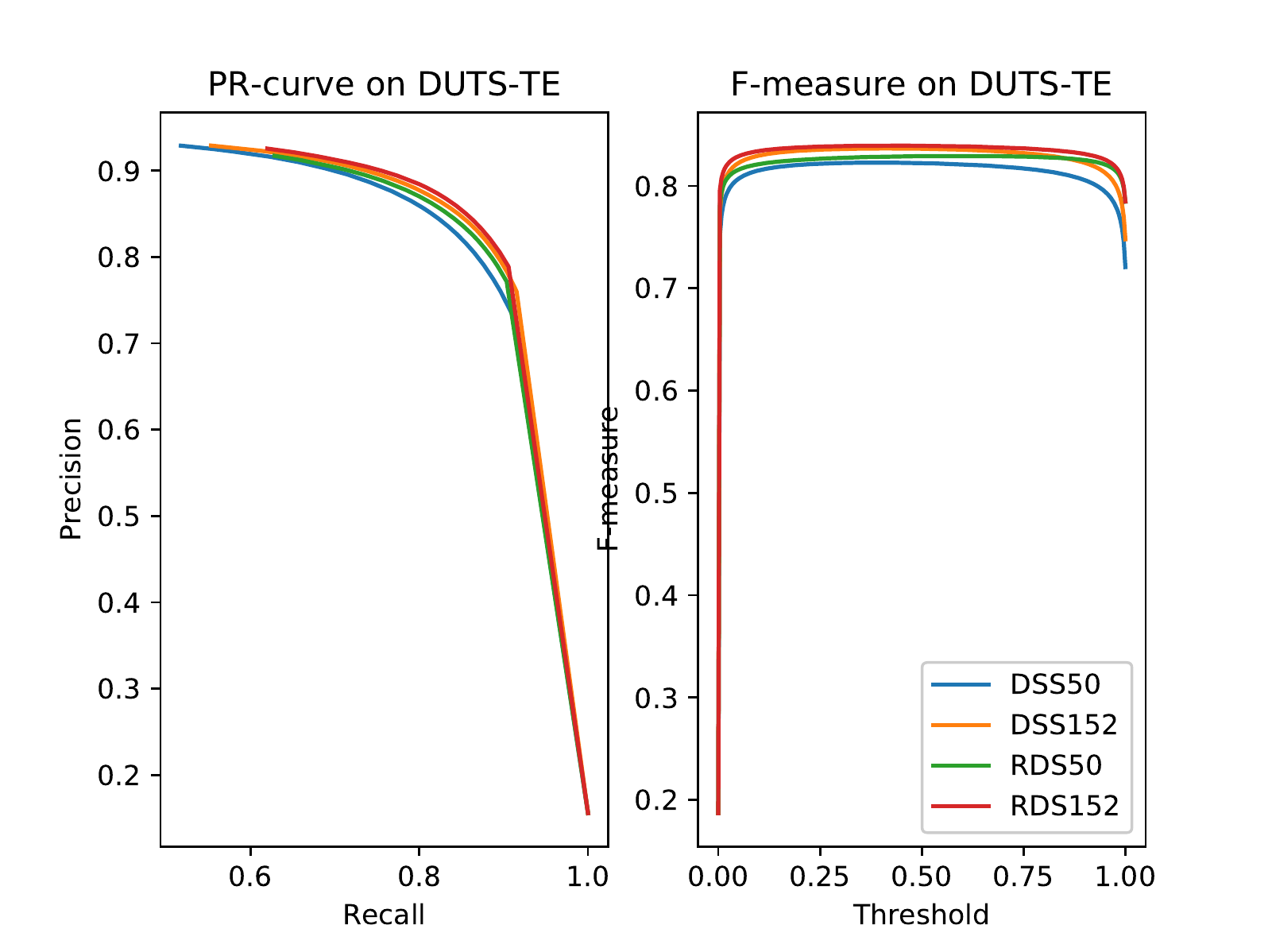}
\end{center}
\caption{PR-curve and F-measure under different thresholds on the five public SOD datasets.}
\label{fig:results1}
\end{figure}

We show the comparison of PR-curve and the F-measure on the five SOD datasets in Figure~\ref{fig:results1}. Generally, the DSS-152 network outperforms DSS-50 due to the deeper architecture. Comparing with DSS-152, our RDS-50 achieved higher results on ECSSD, HKU-IS and PASCAL-S. This shows the effect of more channels can retain richer representation from the top to bottom. The RDS-152 network obtained the best result on most of the benchmarks except DUT-OMRON.

\begin{table*}
\begin{center}
\begin{tabular}{|c|c|c|c|c|c|c|c|c|c|c|c|c|}
\hline
\multirow{2}{*}{Methods} & \multicolumn{2}{c|}{ECSSD}& \multicolumn{2}{c|}{HKU-IS} & \multicolumn{2}{c|}{PASCAL-S}& \multicolumn{2}{c|}{DUT-OMRON}& \multicolumn{2}{c|}{DUTS-TE}&\multicolumn{2}{c|}{PASCAL-SOD} \\ 
  &$F_{\beta}$& $MAE$ &$F_{\beta}$& $MAE$ &$F_{\beta}$& $MAE$ &$F_{\beta}$& $MAE$ &$F_{\beta}$& $MAE$&$F_{\beta}$& $MAE$\\ 
 \hhline{|=|=|=|=|=|=|=|=|=|=|=|=|=|}
 MDF\cite{MDF}  &.881&.105&.897&.129&.836&.150&.769&.091&-&-&.771&.137\\ 
 DCL\cite{DCL}  &.923&.067&.927&.048&-&-&.806&.079&-&-&-&-\\
 ELD\cite{ELD}  &.900&.078&-&-&.828&.125&.765&.091&-&-&.767&.123\\
 DS\cite{DS}  &.915&.121&-&-&.851&.166&.806&.120&-&-&.786&.177\\

 NLDF\cite{NLDF}  &.924&.062&.921&.047&.872&.103&.787&.079&-&-&.799&.105\\
 UCF\cite{UCF}  &.922&.069&.902&.061&.874&.111&.744&.120&.787&.111&.788&.126\\
 AMU\cite{AMU}  &.929&.058&.909&.050&\textcolor{red}{.876}&.098&.761&.097&.789&.084&.795&.112\\
 SRM\cite{SRM}  &.931&.054&.916&.045&\textcolor{red}{.876}&.088&.797&.069&.843&.058&.810&.094\\
 CAR\cite{CAR}  &.937&.050&-&-&.858&.096&.794&.059&-&-&.785&.093\\
 DSS\cite{DSS}  &.919&.051&.914&.040&.842&.104&.769&.063&-&-&.771&.100\\

CKT\cite{CKT}&.928&.054&.910&.048&.875&.085&.793&.072&.829&.062&\textcolor{red}{.818}&.093\\
DGRL\cite{DGRL}  &.917&.040&.905&.035&.868&\textcolor{blue}{.079}&.756&.061&.813&.049&.774&.087\\
LPSD\cite{LPSD}  &.937&.041&.926&.033&.873&\textcolor{blue}{.079}&.780&.060&.840&.048&.792&.086\\
PAGR\cite{PAGR}  &.935&.060&.923&.047&.865&.098&.787&.071&.862&.054&\textcolor{red}{.818}&.095\\

PIC\cite{PIC}  &.942&\textcolor{red}{.034}&.928&.030&.873&\textcolor{red}{.073}&.805&.054&.862&\textcolor{red}{.040}&.805&\textcolor{red}{.078}\\
\hline

RDS-152  &\textcolor{red}{.953}&\textcolor{blue}{.036}&\textcolor{blue}{.942}&\textcolor{red}{.028}&.874&.080&\textcolor{blue}{.837}&\textcolor{blue}{.050}&\textcolor{blue}{.867}&.044&.817&.083\\

RDS-152-OBJ  &\textcolor{red}{.953}&.037&\textcolor{red}{.943}&\textcolor{red}{.028}&.872&.080&\textcolor{red}{.838}&\textcolor{red}{.049}&\textcolor{red}{.871}&\textcolor{blue}{.043}&\textcolor{red}{.818}&\textcolor{blue}{.081}\\
\hline
\end{tabular}
\end{center}
\caption{Quantitative comparison between the RDS network and the state-of-the-art results. The top one is highlighted in red and the second place is in blue.}
\label{tab:state}
\end{table*}

\begin{figure*}[pht]
\begin{center}

\includegraphics[width=2\columnwidth,height=0.2\columnwidth]{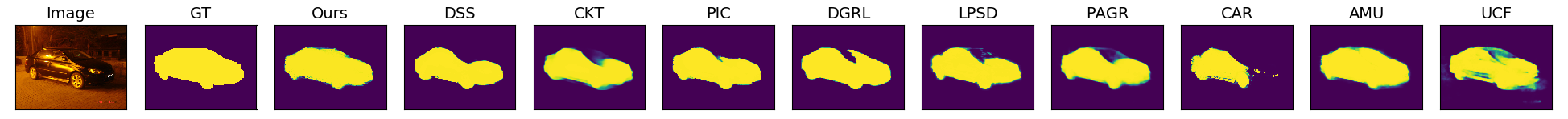}
\includegraphics[width=2\columnwidth,height=0.15\columnwidth]{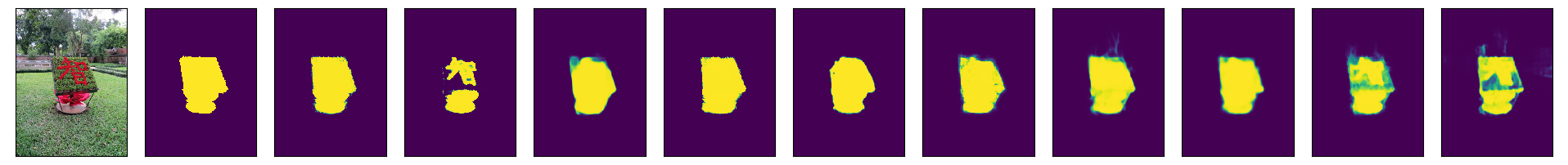}
\includegraphics[width=2\columnwidth,height=0.1\columnwidth]{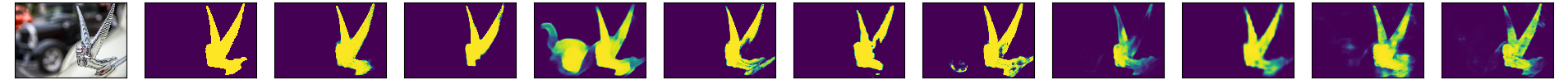}
\includegraphics[width=2\columnwidth,height=0.15\columnwidth]{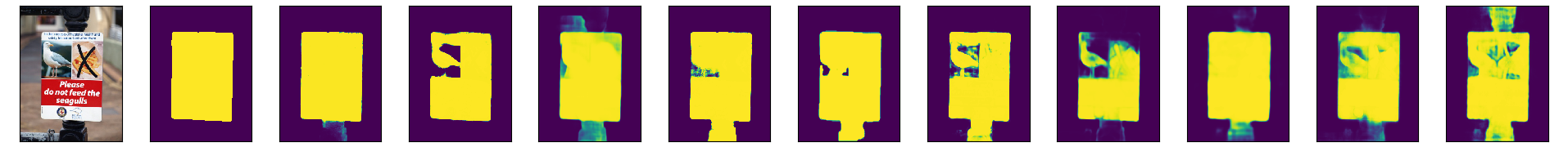}
\includegraphics[width=2\columnwidth,height=0.15\columnwidth]{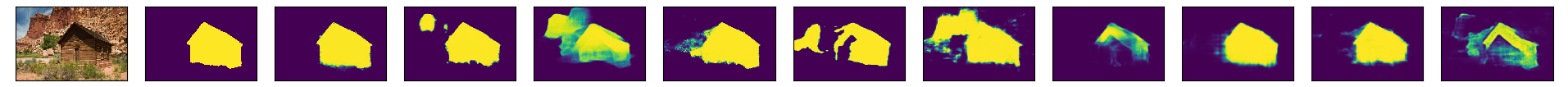}
\includegraphics[width=2\columnwidth,height=0.15\columnwidth]{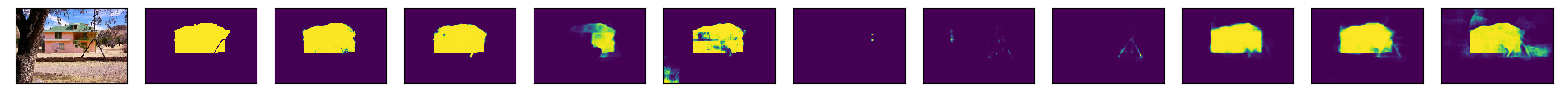}
\includegraphics[width=2\columnwidth,height=0.15\columnwidth]{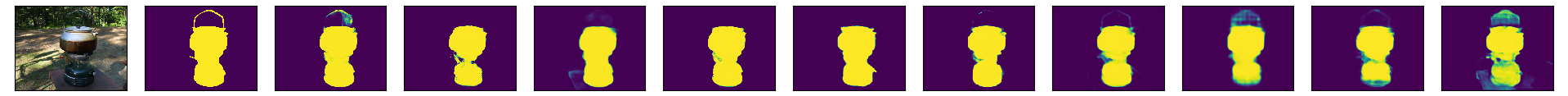}

\end{center}
\caption{Visual comparison between RDS and other state-of-the-art methods. GT: Groundtruth.}
\label{fig:examples}
\end{figure*}

\subsection{RDS Network with Objectness Cues}
In this experiment, we use the setting of RDS-152 from the last experiment to further investigate the effect of objectness cue. The CNN model used for SOD is initialized using the weight that pre-trained on ImageNet. The learned distinguishing object feature can also be applied for SOD even though contextual knowledge is omitted \cite{LPSD}. We train RDS on the merged object dataset to learn coarse location features, but knowledge from ImageNet classification is also desired. Due to the \textit{catastrophic forgetting}, we only fine-tuned the model on the object dataset for one epoch using a learning rate of $0.001$. 
Then the pre-trained network was further fine-tuned on the SOD training set using the same setting from the last experiment.

\begin{figure}[pht]
\begin{center}
\includegraphics[width=1\columnwidth,height=0.45\columnwidth]{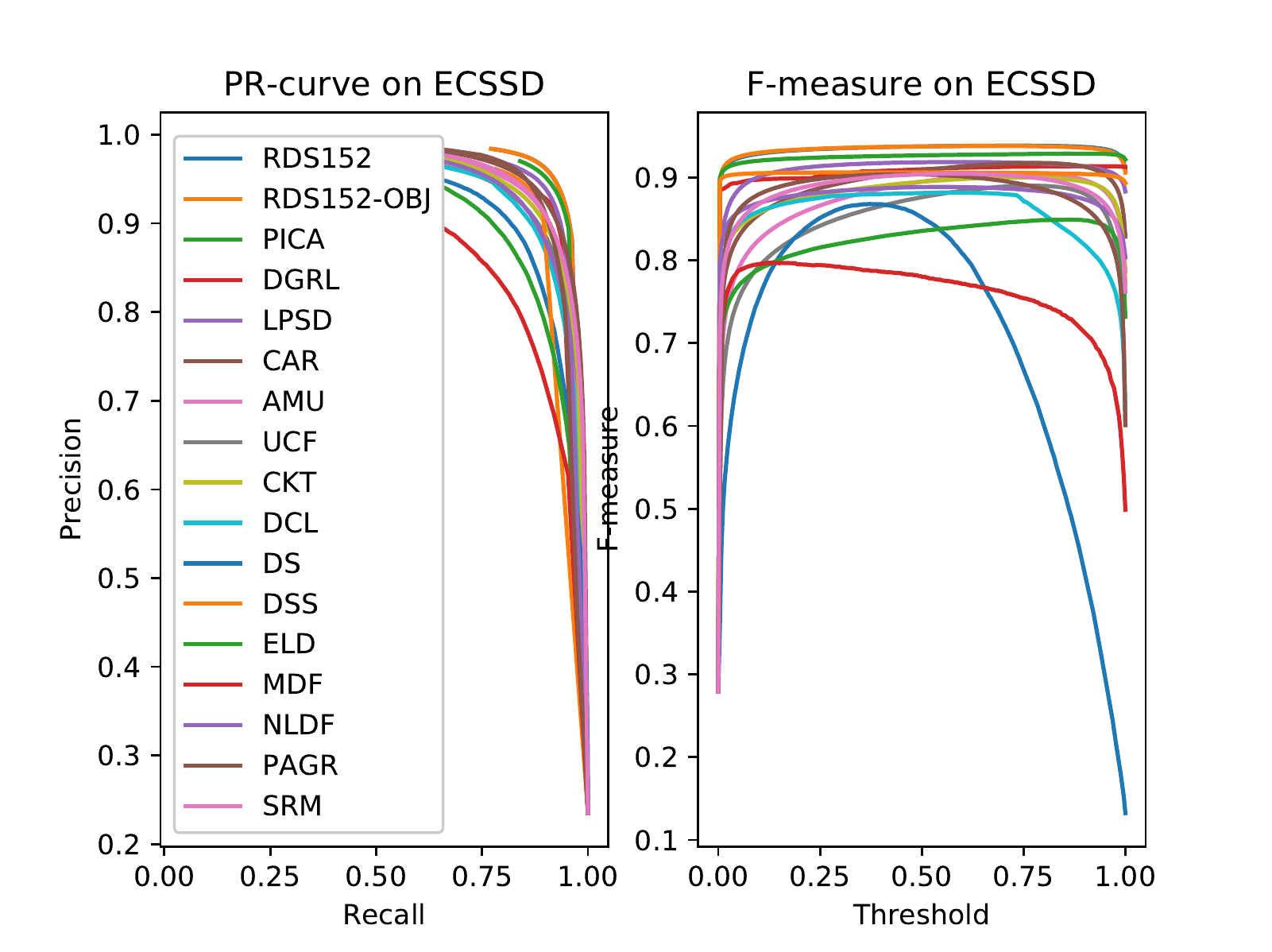}
\includegraphics[width=1\columnwidth,height=0.45\columnwidth]{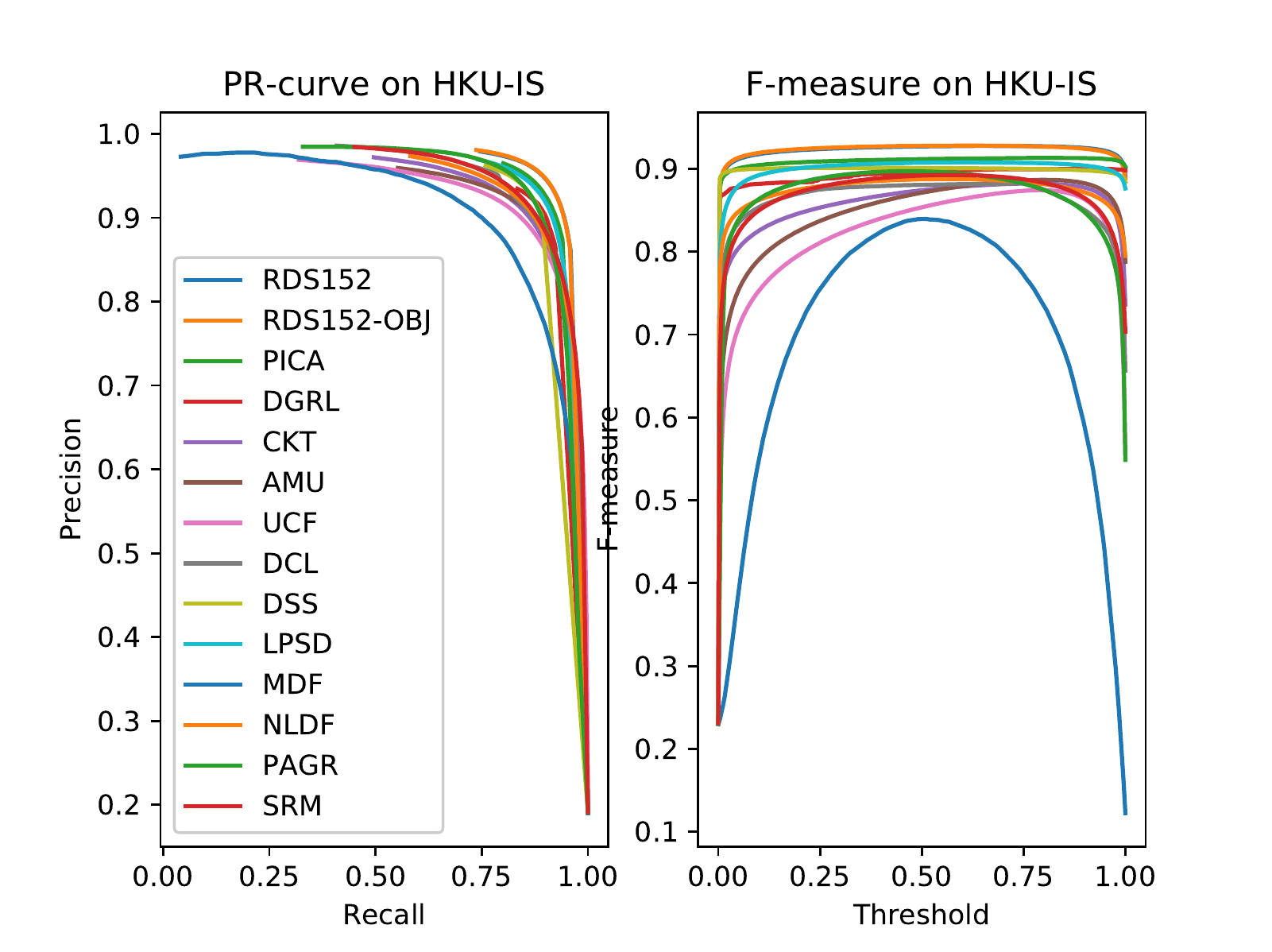}
\includegraphics[width=1\columnwidth,height=0.45\columnwidth]{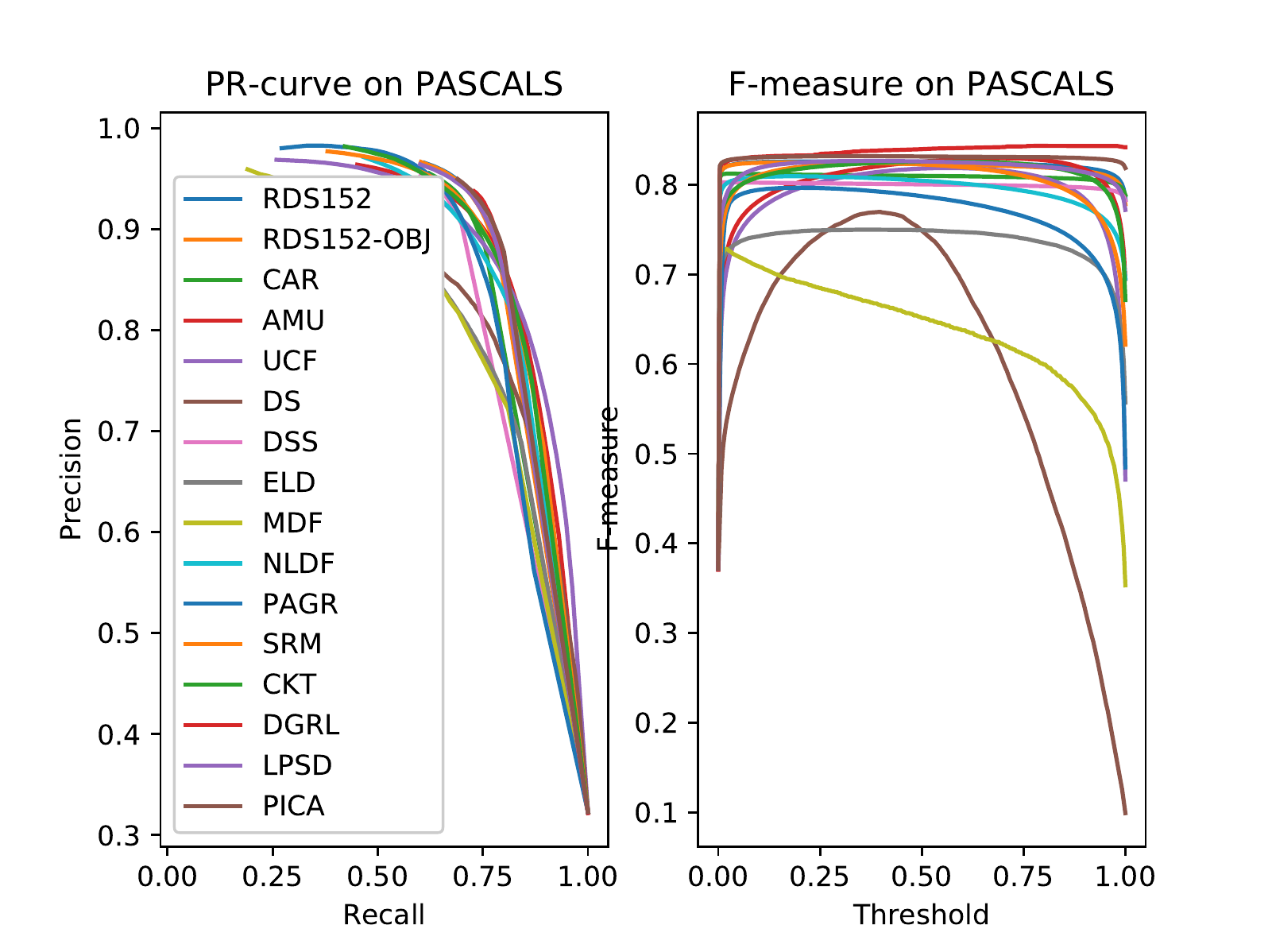}
\includegraphics[width=1\columnwidth,height=0.45\columnwidth]{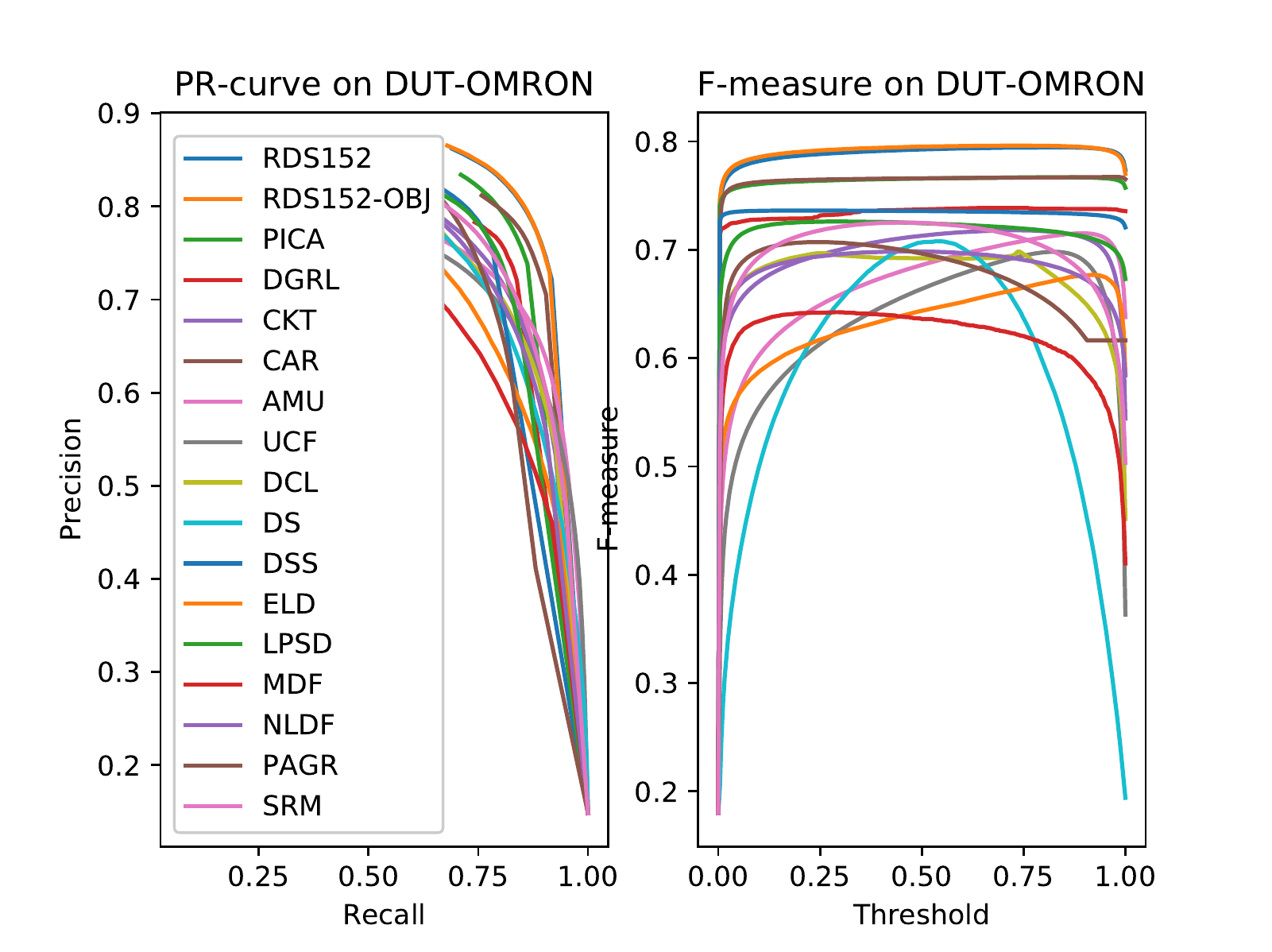}
\includegraphics[width=1\columnwidth,height=0.45\columnwidth]{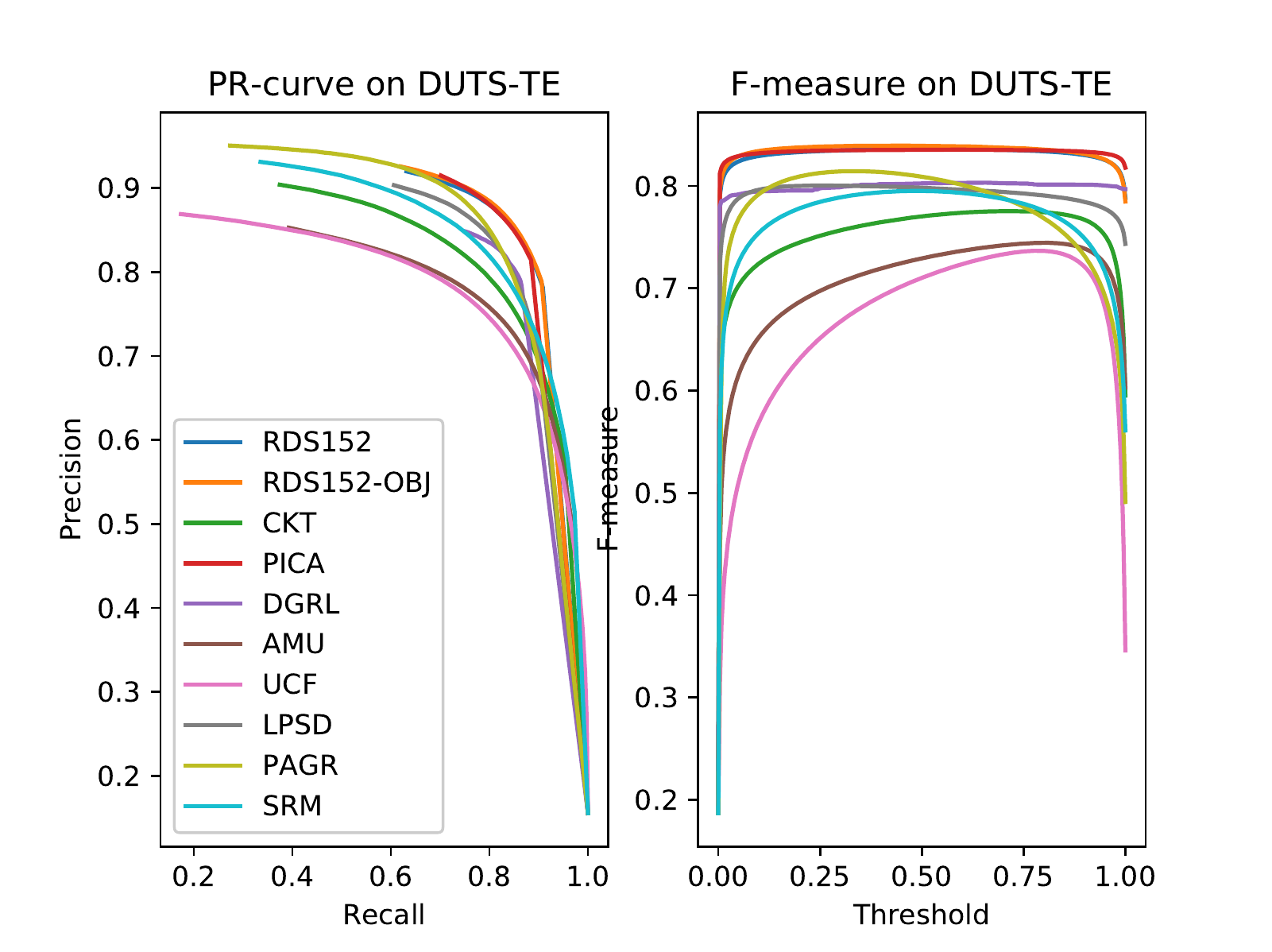}
\end{center}
\caption{PR-curve and F-measure under different thresholds on the five public SOD datasets.}
\label{fig:results2}
\end{figure}

\begin{figure}[pht]
\begin{center}
\includegraphics[width=1\columnwidth,height=0.2\columnwidth]{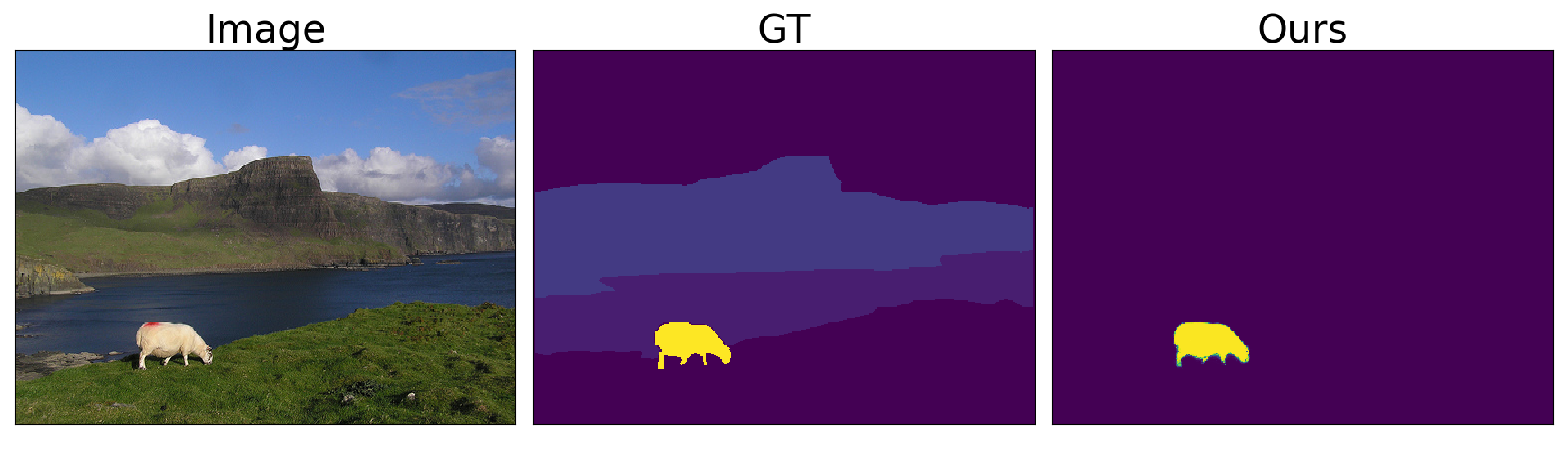}
\includegraphics[width=1\columnwidth,height=0.2\columnwidth]{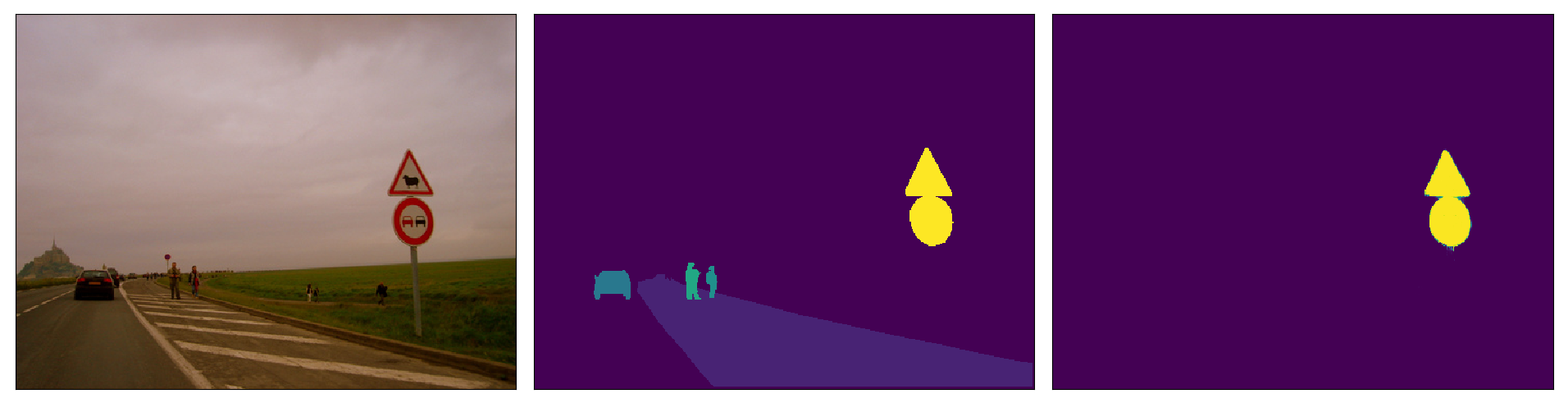}
\includegraphics[width=1\columnwidth,height=0.2\columnwidth]{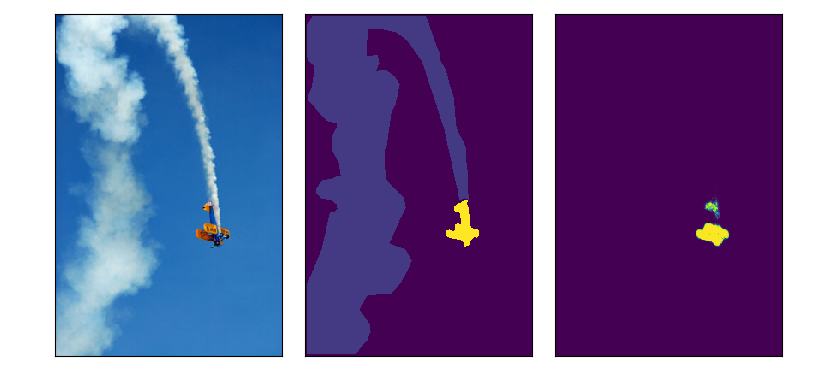}
\end{center}
\caption{Failure case study.}
\label{fig:failure}
\end{figure}

\begin{figure}[pht]
\begin{center}
\includegraphics[width=1\columnwidth,height=0.45\columnwidth]{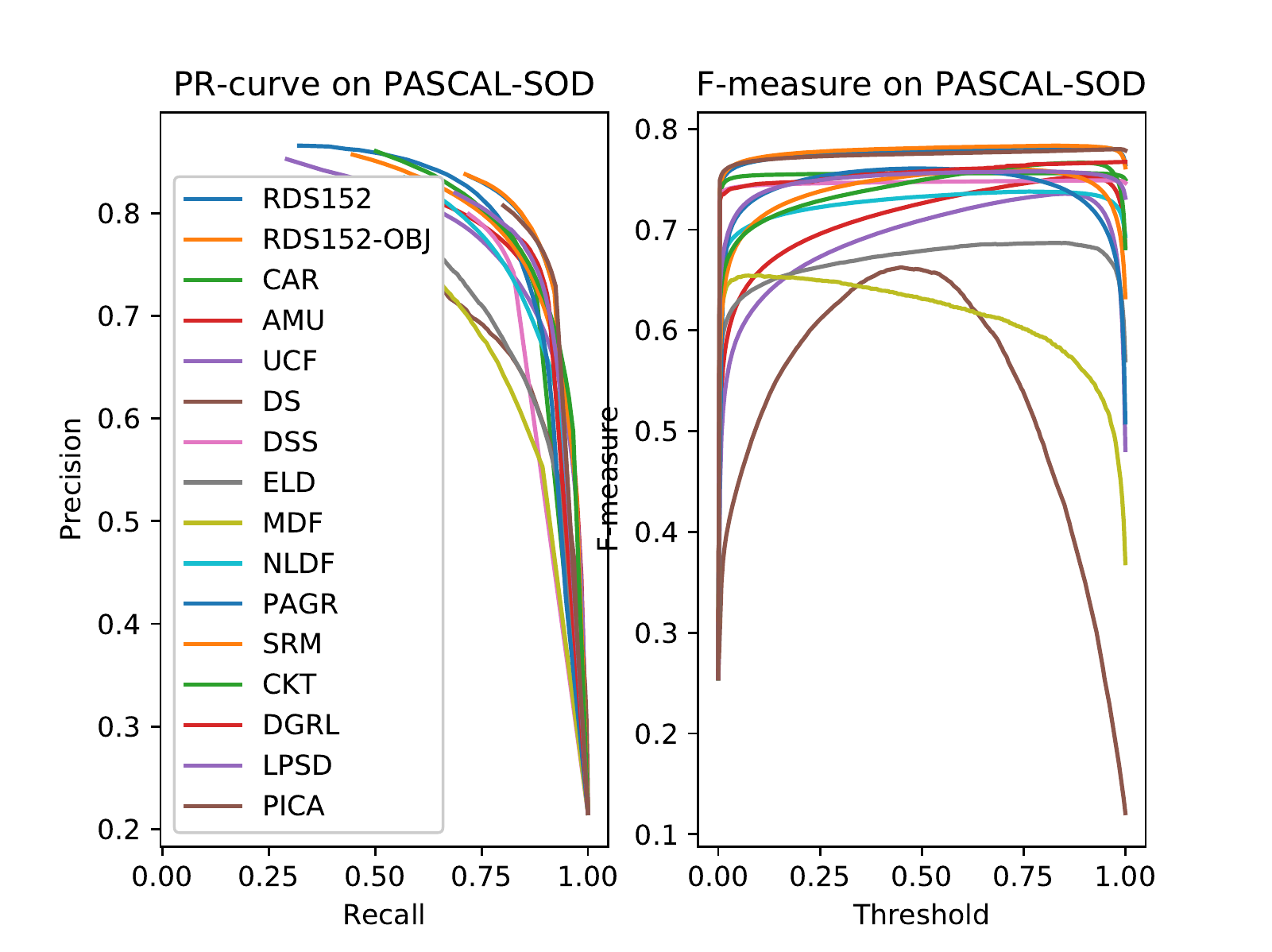}
\end{center}
\caption{PR-curve and F-measure on the PASCAL-SOD dataset.}
\label{fig:results3}
\end{figure}

We compare out methods with $15$ state-of-the-art SOD systems on the five datasets. The saliency maps compared were produced by the authors. In Figure~\ref{fig:results2}, we can see that the RDS-152-OBJ achieved a slightly better result than RDS-152 from the last experiment. But the difference is not large and we believe it is due to the issue of \textit{catastrophic forgetting}. The model weights are learned toward SOD features after a long training process and the prior knowledge fades. Compare with other SOD methods, we achieved the best performance on the DUT-OMRON dataset, where RDS-152-OBJ and RDS-152 were the top two scores. We also obtained the hgihest F-measure on ECSSD, HKU-IS and the DUTS-TE datasets. We show the qualitative comparison between our work and other recently proposed SOD methods in Figure~\ref{fig:examples}. However, our RDS network (RDS-152) achieved $.874$ F-measure on the PASCAL-S dataset, $.002$ lower than AMU \cite{AMU} and SRM \cite{SRM}. When studying the failure case, we found out the ground truth of PASCAL-S may contain background regions with lower pixel values, see Figure~\ref{fig:failure}. In this study, the background is considered as non-salient category during the training phase and detecting background is off-topic. Therefore, we ignore the background region by only considering the most salient object (highest pixel value) to create a new ground truth, denoted as PASCAL-SOD. We show the PR-curve and the F-measure on the PASCAL-SOD in Figure~\ref{fig:results3}. As shown in the last column in Tale~\ref{tab:state}, our RDS-152-OBJ achieved the highest F-measure score, $.818$. 

\section{Conclusion}
In this paper, we have proposed a new structure, RDS, to better combine features of each side output. Richer representations of global and local knowledge can be merged more effectively to deliver a better performance. We have also introduced the images designed for object detection to learn coarse features before fine-tuning the network on the SOD dataset. The result shows the large-scale object data can improve SOD performances. Our proposed network obviously outperforms other competitors by a large margin. But the problem of \textit{catastrophic forgetting} may affect the procedure of pre-training on the object dataset. We will continue study on how to better integrate objectness cues in the future work.



{\small
\bibliographystyle{ieee}
\bibliography{egbib}

\begin{thebibliography}{10}\itemsep=-1pt

\bibitem{Achanta09}
R.~Achanta, S.~Hemami, F.~Estrada, and S.~Susstrunk.
\newblock Frequency-tuned salient region detection.
\newblock In {\em 2009 IEEE Conference on Computer Vision and Pattern
  Recognition}, pages 1597--1604, June 2009.

\bibitem{Alexe10}
B.~Alexe, T.~Deselaers, and V.~Ferrari.
\newblock What is an object?
\newblock In {\em The Twenty-Third {IEEE} Conference on Computer Vision and
  Pattern Recognition, {CVPR} 2010, San Francisco, CA, USA, 13-18 June 2010},
  pages 73--80, 2010.

\bibitem{Survey}
A.~Borji, M.~Cheng, H.~Jiang, and J.~Li.
\newblock Salient object detection: {A} survey.
\newblock {\em CoRR}, abs/1411.5878, 2014.

\bibitem{Borji12}
A.~Borji, S.~Frintrop, D.~N. Sihite, and L.~Itti.
\newblock Adaptive object tracking by learning background context.
\newblock In {\em Computer Vision and Pattern Recognition Workshops (CVPRW),
  2012 IEEE Computer Society Conference on}, pages 23--30. IEEE, 2012.

\bibitem{Cheng15}
M.~Cheng, N.~J. Mitra, X.~Huang, P.~H.~S. Torr, and S.~Hu.
\newblock Global contrast based salient region detection.
\newblock {\em IEEE Transactions on Pattern Analysis and Machine Intelligence},
  37(3):569--582, March 2015.

\bibitem{MSRAK}
M.~Cheng, N.~J. Mitra, X.~Huang, P.~H.~S. Torr, and S.~Hu.
\newblock Global contrast based salient region detection.
\newblock {\em IEEE Transactions on Pattern Analysis and Machine Intelligence},
  37(3):569--582, March 2015.

\bibitem{Cornia18}
M.~Cornia, L.~Baraldi, G.~Serra, and R.~Cucchiara.
\newblock Predicting human eye fixations via an lstm-based saliency attentive
  model.
\newblock {\em IEEE Transactions on Image Processing}, 27(10):5142--5154, Oct
  2018.

\bibitem{ImageNet}
J.~Deng, W.~Dong, R.~Socher, L.~Li, K.~Li, and L.~Fei-Fei.
\newblock Imagenet: A large-scale hierarchical image database.
\newblock In {\em 2009 IEEE Conference on Computer Vision and Pattern
  Recognition}, pages 248--255, June 2009.

\bibitem{PASCALVOC}
M.~Everingham, L.~Van~Gool, C.~K.~I. Williams, J.~Winn, and A.~Zisserman.
\newblock The {PASCAL} {V}isual {O}bject {C}lasses {C}hallenge 2010 {(VOC2010)}
  {R}esults.
\newblock
  http://www.pascal-network.org/challenges/VOC/voc2010/workshop/index.html.

\bibitem{ResNet}
K.~He, X.~Zhang, S.~Ren, and J.~Sun.
\newblock Deep residual learning for image recognition.
\newblock In {\em 2016 IEEE Conference on Computer Vision and Pattern
  Recognition (CVPR)}, pages 770--778, June 2016.

\bibitem{DSS}
Q.~Hou, M.~Cheng, X.~Hu, A.~Borji, Z.~Tu, and P.~H.~S. Torr.
\newblock Deeply supervised salient object detection with short connections.
\newblock {\em IEEE Transactions on Pattern Analysis and Machine Intelligence},
  pages 1--1, 2018.

\bibitem{Hou16}
Q.~Hou, P.~K. Dokania, D.~Massiceti, Y.~Wei, M.-M. Cheng, and P.~H.~S. Torr.
\newblock Mining pixels: Weakly supervised semantic segmentation using image
  labels.
\newblock {\em CoRR}, abs/1612.02101, 2016.

\bibitem{CAR}
M.~A. Islam, M.~Kalash, M.~Rochan, N.~D.~B. Bruce, and Y.~Wang.
\newblock Salient object detection using a context-aware refinement network.
\newblock In {\em BMVC}, 2017.

\bibitem{EML}
S.~Jia.
\newblock {EML-NET:} an expandable multi-layer network for saliency prediction.
\newblock {\em CoRR}, abs/1805.01047, 2018.

\bibitem{Jia16}
S.~Jia, T.~Lansdall-Welfare, and N.~Cristianini.
\newblock Gender classification by deep learning on millions of weakly labelled
  images.
\newblock In {\em 2016 IEEE 16th International Conference on Data Mining
  Workshops (ICDMW)}, pages 462--467, Dec 2016.

\bibitem{Jia13}
Y.~Jia and M.~Han.
\newblock Category-independent object-level saliency detection.
\newblock In {\em 2013 IEEE International Conference on Computer Vision}, pages
  1761--1768, Dec 2013.

\bibitem{Jin17}
B.~Jin, M.~V.~O. Segovia, and S.~Süsstrunk.
\newblock Webly supervised semantic segmentation.
\newblock In {\em 2017 IEEE Conference on Computer Vision and Pattern
  Recognition (CVPR)}, pages 1705--1714, July 2017.

\bibitem{Kim14}
J.~Kim, D.~Han, Y.~Tai, and J.~Kim.
\newblock Salient region detection via high-dimensional color transform.
\newblock In {\em 2014 IEEE Conference on Computer Vision and Pattern
  Recognition}, pages 883--890, June 2014.

\bibitem{crf}
P.~Kr\"{a}henb\"{u}hl and V.~Koltun.
\newblock Efficient inference in fully connected crfs with gaussian edge
  potentials.
\newblock In J.~Shawe-Taylor, R.~S. Zemel, P.~L. Bartlett, F.~Pereira, and
  K.~Q. Weinberger, editors, {\em Advances in Neural Information Processing
  Systems 24}, pages 109--117. Curran Associates, Inc., 2011.

\bibitem{ELD}
G.~Lee, Y.-W. Tai, and J.~Kim.
\newblock Deep saliency with encoded low level distance map and high level
  features.
\newblock {\em 2016 IEEE Conference on Computer Vision and Pattern Recognition
  (CVPR)}, pages 660--668, 2016.

\bibitem{Lei15}
B.~Lei, E.-L. Tan, S.~Chen, D.~Ni, and T.~Wang.
\newblock Saliency-driven image classification method based on histogram mining
  and image score.
\newblock {\em Pattern Recognition}, 48(8):2567 -- 2580, 2015.

\bibitem{MDF}
G.~Li and Y.~Yu.
\newblock Visual saliency based on multiscale deep features.
\newblock In {\em The IEEE Conference on Computer Vision and Pattern
  Recognition (CVPR)}, June 2015.

\bibitem{DCL}
G.~Li and Y.~Yu.
\newblock Deep contrast learning for salient object detection.
\newblock {\em 2016 IEEE Conference on Computer Vision and Pattern Recognition
  (CVPR)}, pages 478--487, 2016.

\bibitem{Li13}
X.~Li, H.~Lu, L.~Zhang, X.~Ruan, and M.~Yang.
\newblock Saliency detection via dense and sparse reconstruction.
\newblock In {\em 2013 IEEE International Conference on Computer Vision}, pages
  2976--2983, Dec 2013.

\bibitem{CKT}
X.~Li, F.~Yang, H.~Cheng, W.~Liu, and D.~Shen.
\newblock Contour knowledge transfer for salient object detection.
\newblock In {\em The European Conference on Computer Vision (ECCV)}, September
  2018.

\bibitem{DS}
X.~Li, L.~Zhao, L.~Wei, M.-H. Yang, F.~Wu, Y.~Zhuang, H.~Ling, and J.~Wang.
\newblock Deepsaliency: Multi-task deep neural network model for salient object
  detection.
\newblock {\em IEEE Transactions on Image Processing}, 25(8):3919 -- 3930, Aug
  2016.

\bibitem{PASCALS}
Y.~Li, X.~Hou, C.~Koch, J.~M. Rehg, and A.~L. Yuille.
\newblock The secrets of salient object segmentation.
\newblock In {\em 2014 IEEE Conference on Computer Vision and Pattern
  Recognition}, pages 280--287, June 2014.

\bibitem{PIC}
N.~Liu, J.~Han, and M.-H. Yang.
\newblock Picanet: Learning pixel-wise contextual attention for saliency
  detection.
\newblock In {\em The IEEE Conference on Computer Vision and Pattern
  Recognition (CVPR)}, June 2018.

\bibitem{MSRA-Liu}
T.~Liu, Z.~Yuan, J.~Sun, J.~Wang, N.~Zheng, X.~Tang, and H.-Y. Shum.
\newblock Learning to detect a salient object.
\newblock {\em IEEE Trans. Pattern Anal. Mach. Intell.}, 33(2):353--367, Feb.
  2011.

\bibitem{NLDF}
Z.~Luo, A.~Mishra, A.~Achkar, J.~Eichel, S.~Li, and P.~Jodoin.
\newblock Non-local deep features for salient object detection.
\newblock In {\em 2017 IEEE Conference on Computer Vision and Pattern
  Recognition (CVPR)}, pages 6593--6601, July 2017.

\bibitem{Mahadevan09}
V.~Mahadevan and N.~Vasconcelos.
\newblock Saliency-based discriminant tracking.
\newblock In {\em 2009 IEEE Conference on Computer Vision and Pattern
  Recognition}, pages 1007--1013, June 2009.

\bibitem{Sharma12}
G.~Sharma, F.~Jurie, and C.~Schmid.
\newblock Discriminative spatial saliency for image classification.
\newblock In {\em 2012 IEEE Conference on Computer Vision and Pattern
  Recognition}, pages 3506--3513, June 2012.

\bibitem{ECSSD}
J.~Shi, Q.~Yan, L.~Xu, and J.~Jia.
\newblock Hierarchical image saliency detection on extended cssd.
\newblock {\em IEEE Transactions on Pattern Analysis and Machine Intelligence},
  38(4):717--729, April 2016.

\bibitem{VGG}
K.~Simonyan and A.~Zisserman.
\newblock Very deep convolutional networks for large-scale image recognition.
\newblock {\em CoRR}, abs/1409.1556, 2014.

\bibitem{Siva13}
P.~Siva, C.~Russell, T.~Xiang, and L.~Agapito.
\newblock Looking beyond the image: Unsupervised learning for object saliency
  and detection.
\newblock In {\em 2013 IEEE Conference on Computer Vision and Pattern
  Recognition}, pages 3238--3245, June 2013.

\bibitem{Sun15}
Y.~Sun, D.~Liang, X.~Wang, and X.~Tang.
\newblock Deepid3: Face recognition with very deep neural networks.
\newblock {\em CoRR}, abs/1502.00873, 2015.

\bibitem{ReNet}
F.~Visin, K.~Kastner, K.~Cho, M.~Matteucci, A.~C. Courville, and Y.~Bengio.
\newblock Renet: A recurrent neural network based alternative to convolutional
  networks.
\newblock {\em CoRR}, abs/1505.00393, 2015.

\bibitem{DUTS}
L.~Wang, H.~Lu, Y.~Wang, M.~Feng, D.~Wang, B.~Yin, and X.~Ruan.
\newblock Learning to detect salient objects with image-level supervision.
\newblock In {\em CVPR}, 2017.

\bibitem{SRM}
T.~Wang, A.~Borji, L.~Zhang, P.~Zhang, and H.~Lu.
\newblock A stagewise refinement model for detecting salient objects in images.
\newblock In {\em 2017 IEEE International Conference on Computer Vision
  (ICCV)}, pages 4039--4048, Oct 2017.

\bibitem{DGRL}
T.~Wang, L.~Zhang, S.~Wang, H.~Lu, G.~Yang, X.~Ruan, and A.~Borji.
\newblock Detect globally, refine locally: A novel approach to saliency
  detection.
\newblock In {\em The IEEE Conference on Computer Vision and Pattern
  Recognition (CVPR)}, June 2018.

\bibitem{SUN}
J.~Xiao, J.~Hays, K.~A. Ehinger, A.~Oliva, and A.~Torralba.
\newblock Sun database: Large-scale scene recognition from abbey to zoo.
\newblock In {\em 2010 IEEE Computer Society Conference on Computer Vision and
  Pattern Recognition}, pages 3485--3492, June 2010.

\bibitem{HED}
S.~Xie and Z.~Tu.
\newblock Holistically-nested edge detection.
\newblock {\em 2015 IEEE International Conference on Computer Vision (ICCV)},
  pages 1395--1403, 2015.

\bibitem{OMRON}
C.~Yang, L.~Zhang, R.~X. Lu, Huchuan, and M.-H. Yang.
\newblock Saliency detection via graph-based manifold ranking.
\newblock In {\em Computer Vision and Pattern Recognition (CVPR), 2013 IEEE
  Conference on}, pages 3166--3173. IEEE, 2013.

\bibitem{LPSD}
Y.~Zeng, H.~Lu, L.~Zhang, M.~Feng, and A.~Borji.
\newblock Learning to promote saliency detectors.
\newblock In {\em The IEEE Conference on Computer Vision and Pattern
  Recognition (CVPR)}, June 2018.

\bibitem{AMU}
P.~Zhang, D.~Wang, H.~Lu, H.~Wang, and X.~Ruan.
\newblock Amulet: Aggregating multi-level convolutional features for salient
  object detection.
\newblock {\em 2017 IEEE International Conference on Computer Vision (ICCV)},
  pages 202--211, 2017.

\bibitem{UCF}
P.~Zhang, D.~Wang, H.~Lu, H.~Wang, and B.~Yin.
\newblock Learning uncertain convolutional features for accurate saliency
  detection.
\newblock {\em 2017 IEEE International Conference on Computer Vision (ICCV)},
  pages 212--221, 2017.

\bibitem{PAGR}
X.~Zhang, T.~Wang, J.~Qi, H.~Lu, and G.~Wang.
\newblock Progressive attention guided recurrent network for salient object
  detection.
\newblock In {\em The IEEE Conference on Computer Vision and Pattern
  Recognition (CVPR)}, June 2018.

\end{thebibliography}
}

\end{document}